\documentclass[]{onethree}

\AtBeginDocument{%
  \providecommand\BibTeX{{%
    Bib\TeX}}}

\usepackage{float}
\usepackage{enumitem}
\usepackage{colortbl}
\usepackage{amssymb}
\usepackage{tabularx}
\usepackage{longtable}
\providecommand{\Description}[1]{}

\newcommand{\wenjia}[1]{\textcolor{blue}{[Wenjia: #1]}}
\newcommand{\wenjiaadd}[1]{\textcolor{blue}{ #1}}

\newcommand{\zhiwei}[1]{\textcolor{red}{[zhiwei: #1]}}
\newcommand{\zhiweiadd}[1]{\textcolor{red}{ #1}}

\newcommand{\pjc}[1]{\textcolor{red}{#1}}

\newcommand{\choucisan}[1]{\textcolor{purple}{#1}}

\begin{document}

\title{Self in Space: Benchmarking Self-Awareness and Spatial Cognition in UAV Embodied Intelligence}

\author[1]{\href{https://choucisan.github.io}{\textcolor{black}{Zhishan Zou}}}
\author[1]{\href{https://github.com/sunguoyan17-alt}{\textcolor{black}{Guoyan Sun}}}
\author[2]{\href{https://trentonwei.github.io}{\textcolor{black}{Zhiwei Wei}}}
\author[3]{\href{https://jianchengpan.space}{\textcolor{black}{Jiancheng Pan}}}
\author[1]{\href{https://github.com/Davidup1}{\textcolor{black}{Yujie Li}}}
\author[1]{\href{https://teacher.bupt.edu.cn/pengmugen/zh_CN/index.htm}{\textcolor{black}{Mugen Peng}}}
\author[1\dagger]{\href{https://teacher.bupt.edu.cn/xuwenjia/zh_CN/index.htm}{\textcolor{black}{Wenjia Xu}}}

\affiliation[1]{Beijing University of Posts and Telecommunications}
\affiliation[2]{Hunan Normal University}
\expandafter\def\expandafter\affiliationlist\expandafter{\affiliationlist\\\affiliationformat[3]{Tsinghua University}}



\abstract{
Autonomous UAV systems increasingly rely on multimodal large language models (MLLMs) in complex real-world environments, requiring coherent representations of both the surrounding space and the agent itself. However, existing UAV-oriented approaches and benchmarks remain largely environment-centric, primarily focusing on spatial understanding tasks, with the agent's self-awareness remaining implicit. To address this gap, we introduce \textbf{SIS-Bench}, a benchmark for evaluating embodied spatial intelligence in UAV scenarios under a unified self-in-space formulation. SIS-Bench organizes evaluation along two complementary dimensions, \emph{space} and \emph{self}, and a three-level hierarchy of perception, memory, and reasoning. It contains 4,856 question--answer pairs across 13 tasks derived from 1,646 real-world UAV videos through a task-conditioned construction pipeline with expert verification.
Extensive evaluations reveal a clear imbalance between spatial cognition and self-awareness in current MLLMs, together with progressive performance degradation across cognitive levels.
Motivated by these findings, we further explore a motion-aware representation that incorporates self-related dynamics through optical flow and visual feature fusion. Experimental results show that modeling agent motion consistently improves perception and memory performance, not only in spatial cognition but also in self-awareness, and generalizes to downstream UAV decision-making tasks.
Our results highlight the importance of self-awareness for advancing embodied spatial intelligence, and provide both a new benchmark and empirical evidence for motion-aware self-in-space modeling.

}


\checkdata[More resources]{
  \href{mailto:choucisan@gmail.com}{\raisebox{-1ex}{\includegraphics[height=4.2ex]{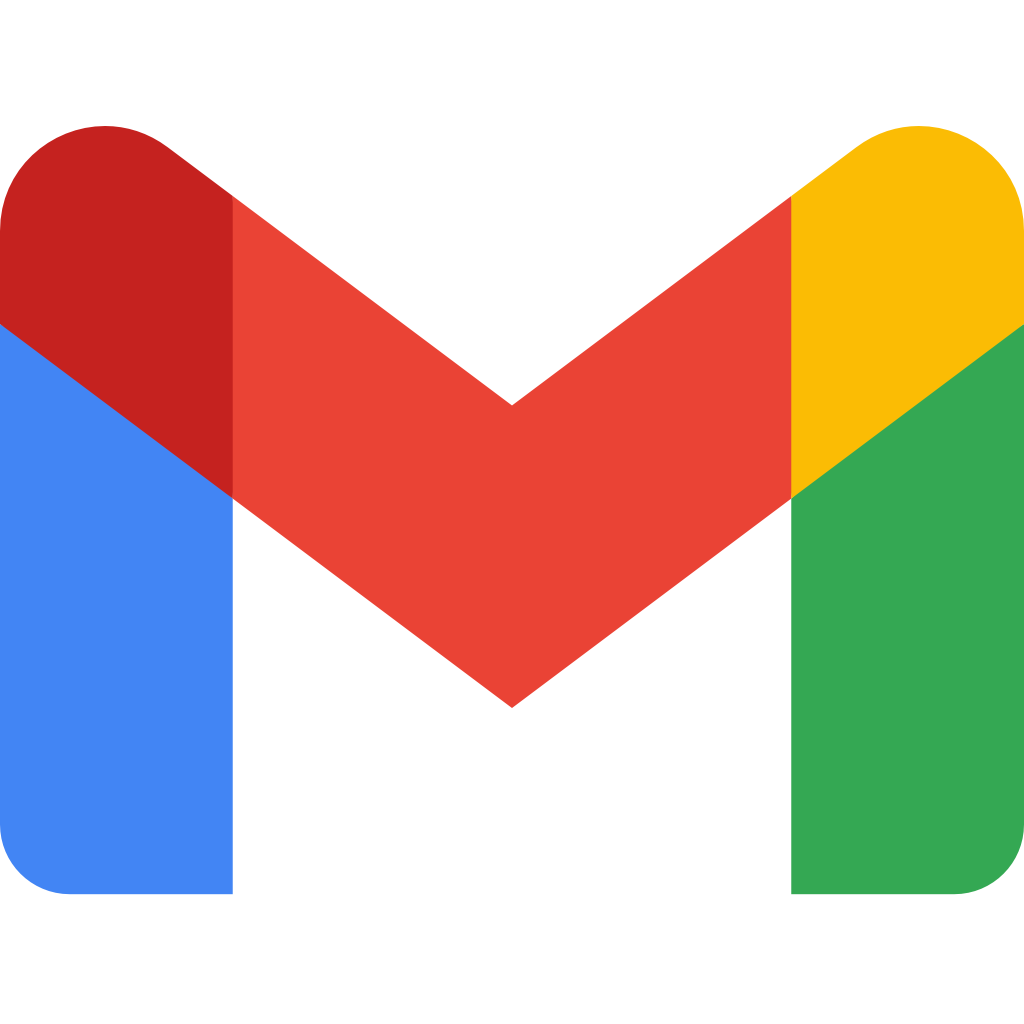}}~Contact}\hspace{1.5em}
  \href{https://choucisan.github.io/publications/self-in-space}{\raisebox{-1ex}{\includegraphics[height=4.2ex]{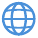}}~Website}\hspace{1.5em}
  \href{https://github.com/IntelliSensing/Self-in-Space}{\raisebox{-1ex}{\includegraphics[height=4.2ex]{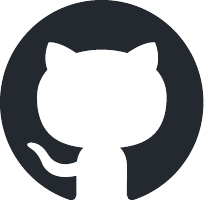}}~Code}\hspace{1.5em}
  \href{https://huggingface.co/collections/choucsan/self-in-space}{\raisebox{-1ex}{\includegraphics[height=4.2ex]{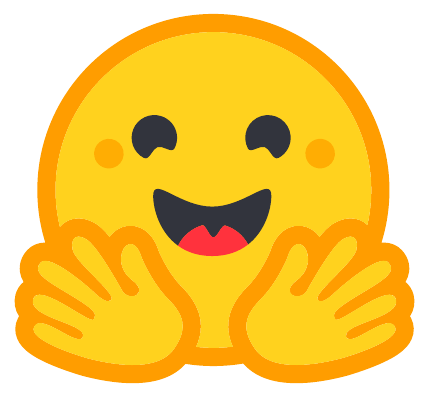}}~Collection}\hspace{1.5em}
  \href{https://www.modelscope.cn/collections/choucisan/Self-in-Space}{\raisebox{-1ex}{\includegraphics[height=4.2ex]{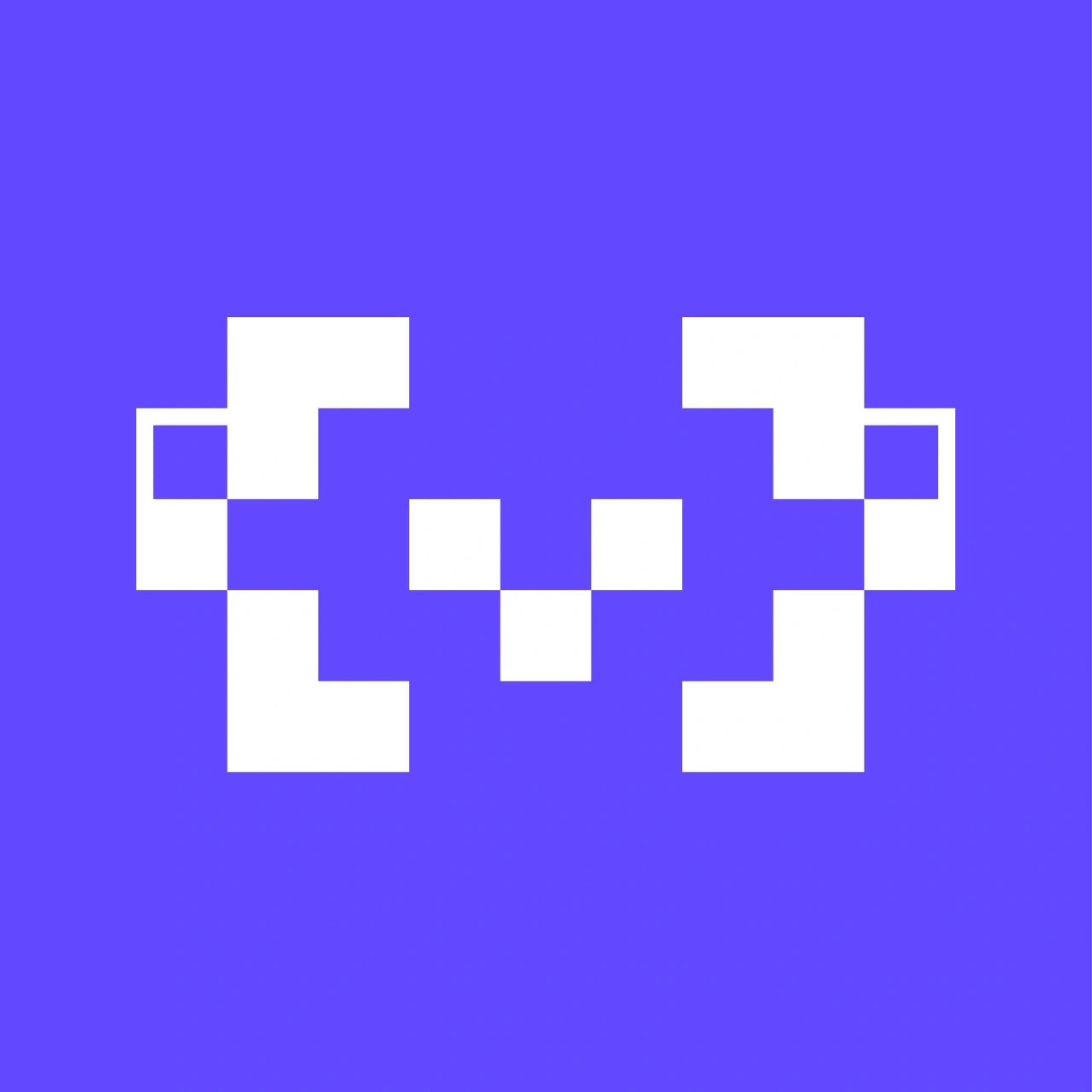}}~Collection}\hspace{1.5em}
}

\vspace*{-6mm}
\maketitle
\vspace{-8mm}

\begin{figure}[h]
  \centering
  \includegraphics[width=1\linewidth]{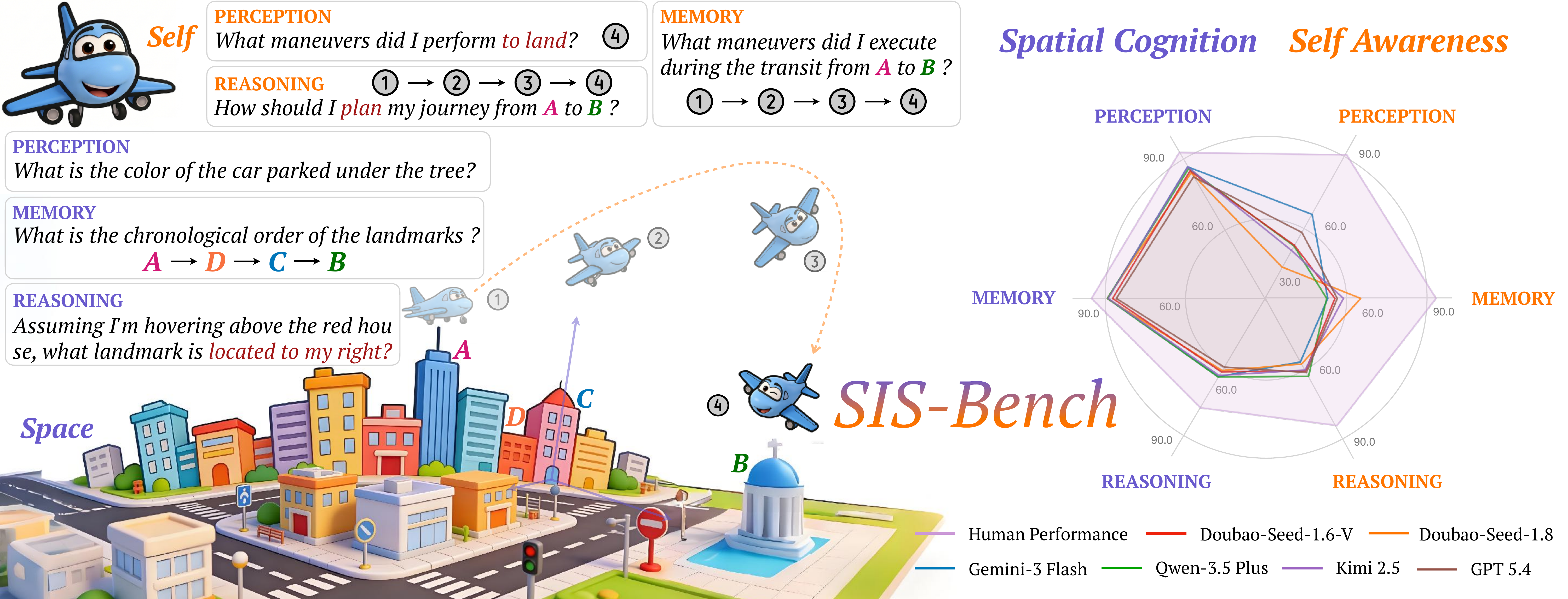}
  \caption{Overview of SIS-Bench. Representative evaluation tasks are
illustrated on the left. The right compares human performance with six
proprietary models, where the purple-labeled axes represent spatial
cognition tasks and the orange-labeled axes represent self-awareness
tasks across perception, memory, and reasoning.}
  \label{fig:teaser}
\end{figure}

\newpage
\tableofcontents
\newpage

\section{Introduction}
\label{sec:intro}

Driven by advances in lightweight hardware, battery technology, and manufacturing, unmanned aerial vehicles (UAVs) have become increasingly available, leading to their widespread deployment in real-world applications such as urban monitoring~\cite{UrbanVideo-Bench}, infrastructure inspection~\cite{aela2024uav}, and emergency response~\cite{ishiwatari2024leveraging}. As these applications grow in scale and complexity, UAV systems must operate more intelligently in dynamic physical environments, requiring stronger capabilities for perception, reasoning, and decision-making~\cite{wen2019visdrone}. Recent advances in multimodal large language models (MLLMs), with their ability to unify perception and reasoning, have thus created new opportunities for UAV intelligence~\cite{UAV-VLN,emami2026large}.

Recent efforts have pushed the boundaries of UAV intelligence with MLLMs, extending their use from aerial perception~\cite{liu2024remoteclip,wang2024skyscript} to navigation~\cite{wang2024towards,xu2025aerial}, and more recently to embodied UAV tasks~\cite{gao2025openfly,liu2026indooruav}. Alongside these advances, a parallel line of work has increasingly recognized the limitations of current MLLMs in UAV scenarios, motivating benchmarks and evaluation frameworks that assess their capabilities, failure modes, and reasoning gaps from different perspectives~\cite{UrbanVideo-Bench,UAVScenes,MM-UAVBench}. Despite these advances, existing studies remain largely environment-centered and task-oriented, which mainly focus on how UAVs perceive the environment and accomplish predefined tasks, while largely overlooking the explicit modeling of the internal state of the UAV itself as an embodied agent. In practice, however, UAV operation is a continuous agent--environment interaction process, in which the UAV is not merely an observer of space, but an active entity evolving within it~\cite{zhang2025mllms,montello1993scale}. Such a setting requires a unified capability that jointly models the external environment (\emph{space}) and the agent’s own state (\emph{self}).

Viewed from this perspective, embodied intelligence is not only about understanding the world, but also about understanding the agent within the world---what we refer to as \textbf{self in space}.

Recent research has increasingly highlighted the importance of agent-centered representations in embodied intelligence, suggesting that explicitly modeling internal agent states and dynamics is critical for grounded and consistent behavior~\cite{DeconstructingSpatialIntelligence,kadambi2025embodiment,li2024embodied}. However, current UAV-oriented MLLMs still fall short of a unified understanding of \emph{space} and \emph{self}. This naturally raises a central question: \textbf{how well do MLLMs jointly model the external environment, the UAV self-state, and the interaction between them?}

To systematically answer this question, we introduce \textbf{SIS-Bench} (\textbf{S}elf-\textbf{I}n-\textbf{S}pace Benchmark), a benchmark for evaluating how well MLLMs jointly model space, self, and their interaction in UAV scenarios. Unlike existing evaluations that mainly focus on environment understanding or task completion, SIS-Bench is built upon a unified self-in-space framework that evaluates UAV embodied intelligence along two complementary dimensions, \emph{spatial cognition} and \emph{self-awareness}, while organizing tasks into a three-level cognitive hierarchy of perception, memory, and reasoning. We further establish SIS-Bench through a task-conditioned construction protocol with heterogeneous video types, task-specific annotation pipelines, and rigorous dual-expert verification.
SIS-Bench contains 4,856 question--answer (QA) pairs across 13 tasks derived from real-world aerial videos spanning diverse environments (e.g., urban, residential, industrial, and natural) and long temporal durations from 10 seconds to over 2 minutes.
Using SIS-Bench, we evaluate selected proprietary and open-source MLLMs and include human study as an upper-bound reference. As shown in Figure~\ref{fig:teaser}, we observe a clear imbalance between spatial cognition and self-awareness, as well as a steady decline across perception, memory, and reasoning, highlighting that existing models remain largely environment-centric and struggle with embodied self-awareness.

Motivated by these findings, we further examine whether explicitly strengthening the joint modeling of \emph{space} and \emph{self} can lead to measurable gains. Rather than presenting a new flagship model, we use this question to conduct a controlled motion-aware exploration. Concretely, we construct \textbf{SIS-Motion}, a motion-aware extension of a standard video MLLM that fuses optical-flow-based motion features with visual embeddings, enabling the model to jointly capture environmental context and agent dynamics. To support this exploration, we construct a spatial motion-aware question answering dataset, \textbf{SIS-Motion-54K}, and perform supervised fine-tuning. Experimental results show that this exploratory setup consistently improves both spatial cognition and self-awareness on SIS-Bench. Moreover, the resulting gains transfer beyond benchmark settings to downstream UAV navigation decision-making tasks~\cite{anderson2018vision,fried2018speaker,irshad2021hierarchical}. Our main contributions are as follows:

\begin{itemize}
    \item We introduce \textbf{SIS-Bench}, a benchmark with 4,856 QA pairs from 1,646 real-world UAV videos, covering 13 tasks over spatial cognition and self-awareness under a perception--memory--reasoning hierarchy.
    \item We evaluate 26 video MLLMs, including 6 proprietary and 20 open-source models, and reveal two consistent limitations: weaker modeling of \emph{self} than \emph{space}, and progressive degradation from perception to memory to reasoning.
    \item Motivated by these findings, we conduct a controlled motion-aware exploration through \textbf{SIS-Motion}, showing improved spatial cognition and self-awareness on SIS-Bench, with transfer to downstream UAV navigation tasks.
\end{itemize}

\section{Related Work}
\label{sec:related}

\noindent\textbf{MLLMs for UAV Applications.}
Recent advances in multimodal large language models (MLLMs) have driven growing interest in UAV-oriented multimodal intelligence. Existing work has progressed from aerial visual understanding, such as image understanding, object detection, and change detection~\cite{liu2024remoteclip,wang2024skyscript}, to navigation and decision-making, where MLLMs support trajectory planning, target-oriented navigation, and spatio-temporal scene interpretation~\cite{wang2024towards,xu2025aerial}. More recent studies further extend UAV intelligence toward agentic and embodied settings, emphasizing autonomous reasoning and closed-loop interaction with the physical world~\cite{koubaa2025agentic,emami2026large,gao2025openfly,liu2026indooruav}. In parallel, dedicated UAV benchmarks have emerged to evaluate capabilities such as perception, navigation, reasoning, and planning under realistic aerial conditions~\cite{UrbanVideo-Bench,UAVScenes,MM-UAVBench}. Unlike these task-centric efforts, our work focuses on how well MLLMs jointly model the external environment, the UAV self-state, and their interaction.

\noindent\textbf{MLLMs for Spatial Intelligence.}
Spatial intelligence enables embodied agents to perceive, represent, and reason about spatial structure in the physical world. Prior work has evolved from vision--language alignment~\cite{radford2021learning,li2023blip} to 3D geometric reasoning and structured spatial grounding~\cite{chen2024ll3da,hong20233d,chen2024spatialvlm,chen2024grounded,wang2023chat,huang2023embodied,huang2024chat,fu2024scene,zhu2024llava,zheng2025video,qi2025gpt4scene}, and more recently to embodied understanding in egocentric and interactive environments~\cite{EgoSchema,majumdar2024openeqa,fu2025objectrelator,pan2026v,mahdi2025exo2egosyn}. While these advances substantially improve spatial reasoning, most existing methods remain environment-centered, with limited explicit modeling of the agent itself. For embodied UAVs, however, robust operation requires consistent representations of self-state, motion, and agent--environment interaction over time. Our work builds on this gap through the joint modeling of \emph{space}, \emph{self}, and their interaction.

\noindent\textbf{Motion-aware Modeling for Embodied UAVs.} Most video-based MLLMs inherit CLIP-style pretraining, which captures high-level semantics but is less effective for fine-grained motion and temporal dynamics. Recent work adds structured spatial cues and embodied video reasoning to improve spatio-temporal understanding~\cite{chen2024spatialvlm,zheng2025video,majumdar2024openeqa}, yet explicit modeling of agent-related motion remains underexplored, especially in UAV scenarios where viewpoint change and self-motion are fundamental. This gap motivates our motion-aware exploration in UAV settings, where we examine whether integrating visual appearance with motion-aware cues can improve self-in-space modeling.

\section{SIS-Bench}
\label{sec:SIS-Bench}

\noindent To systematically study UAV embodied spatial intelligence, we introduce \textbf{SIS-Bench}, a benchmark that decomposes this capability into structured and measurable components. Rather than evaluating isolated perception or reasoning skills, SIS-Bench is designed to capture how a UAV jointly models the external environment and its own evolving action state under realistic embodied scenarios. Following this formulation, we first present the design principles in Sec.~\ref{sec:design}, then describe the benchmark construction pipeline in Sec.~\ref{sec:pipeline}, and finally summarize dataset statistics in Sec.~\ref{sec:stats}.

\begin{figure*}[t]
	\centering
    \vspace{-10pt}
	\includegraphics[width = 1\linewidth]{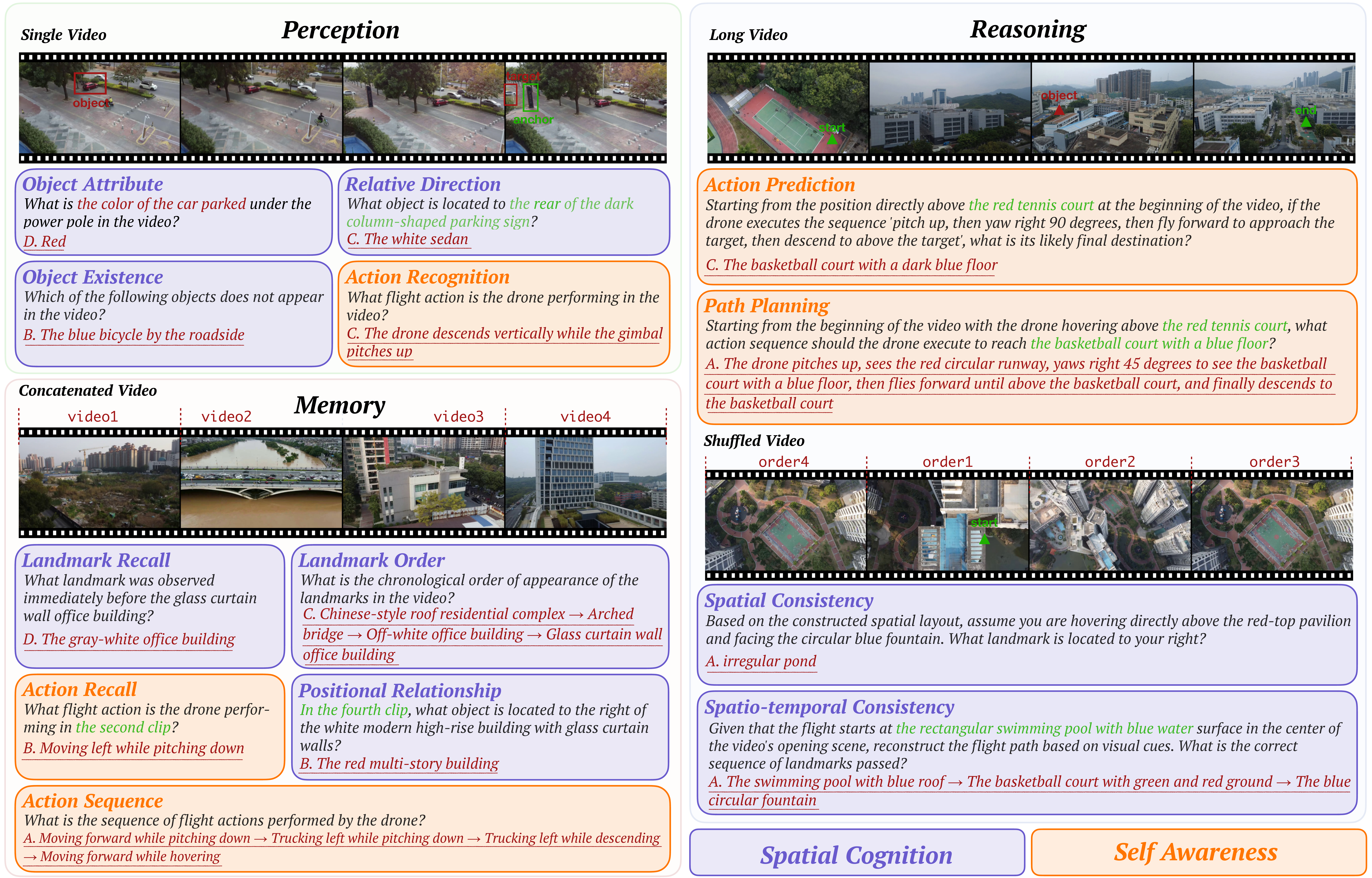}
	\vspace{-10pt}
        \caption{Task illustrations of SIS-Bench. The benchmark evaluates 13 tasks under two dimensions (Spatial Cognition and Self-Awareness) and three cognitive levels. Perception is built from \emph{Single Video} inputs, Memory from \emph{Concatenated Video} inputs, and Reasoning from \emph{Long Video} and \emph{Shuffled Video} inputs, reflecting progressively higher cognitive demands. Colored boxes are for illustration only and are not part of the model input.}
	\label{Fig:task}
	\vspace{-10pt}
\end{figure*}

\subsection{Design Principles}
\label{sec:design}

Unlike existing benchmarks that mainly focus on environment understanding or downstream task completion, SIS-Bench is built around a \textbf{self-in-space} formulation. The goal is to evaluate not only whether a model understands the surrounding scene, but also whether it can maintain a coherent representation of the UAV itself as an embodied agent acting within that scene.

\noindent\textbf{Self-in-Space Dual-Dimension.}
As shown in Figure~\ref{Fig:task}, we organize the benchmark along two complementary dimensions. (1) \textbf{Spatial cognition} measures how well a model understands the external environment, including objects, landmarks, spatial relations, and scene consistency. (2) \textbf{Self-awareness} measures how well it understands the UAV's own motion, action history, and future behavior. This dual design directly follows the central question in Sec.~\ref{sec:intro}: embodied UAV intelligence requires joint modeling of \emph{space}, \emph{self}, and their interaction.

\noindent\textbf{Hierarchical Cognitive Design.}
Within each of these two dimensions, we further organize the benchmark into three progressive cognitive levels: \textbf{perception}, \textbf{memory}, and \textbf{reasoning}.
Inspired by human cognition, this hierarchy captures the evolution from immediate observation to temporal retention and structured inference, enabling a more fine-grained evaluation than a single aggregate score.
\textbf{(1) Perception.}
Perception focuses on immediate understanding of visual scenes and agent actions. Under spatial cognition, it includes \textit{Object Existence}, \textit{Object Attribute}, and \textit{Relative Direction}. Under self-awareness, it includes \textit{Action Recognition}. This level isolates instantaneous understanding without requiring temporal accumulation.
\textbf{(2) Memory.}
Memory evaluates whether a model can retain and retrieve information over time. Under spatial cognition, it includes \textit{Landmark Recall}, \textit{Landmark Order}, and \textit{Positional Relationship}. Under self-awareness, it includes \textit{Action Sequence} and \textit{Action Recall}. This level emphasizes temporal continuity and information persistence beyond immediate observation.
\textbf{(3) Reasoning.}
Reasoning evaluates higher-level embodied inference that integrates spatial context and agent dynamics. Under spatial cognition, it includes \textit{Spatial Consistency} and \textit{Spatio-temporal Consistency}. Under self-awareness, it includes \textit{Action Prediction} and \textit{Path Planning}. This level places the strongest demands on structured reasoning across both \emph{space} and \emph{self}.

\noindent\textbf{Task-conditioned Video Construction.}
To match task demands, we define four video construction types. (1) \textbf{Single Video} keeps one self-contained observation for instantaneous perception. (2) \textbf{Concatenated Video} composes 2--4 \textbf{Single Video} clips to add cross-segment dependency for memory evaluation. (3) \textbf{Long Video} preserves long-horizon motion and scene evolution for future-action reasoning. (4) \textbf{Shuffled Video} permutes segmented \textbf{Long Video} clips to break original chronology, requiring recovery of spatial and temporal consistency beyond the input order. Based on this design, SIS-Bench prioritizes task-driven data over uniform pooling.

\begin{figure*}[t]
	\centering
	\includegraphics[width = 1\linewidth]{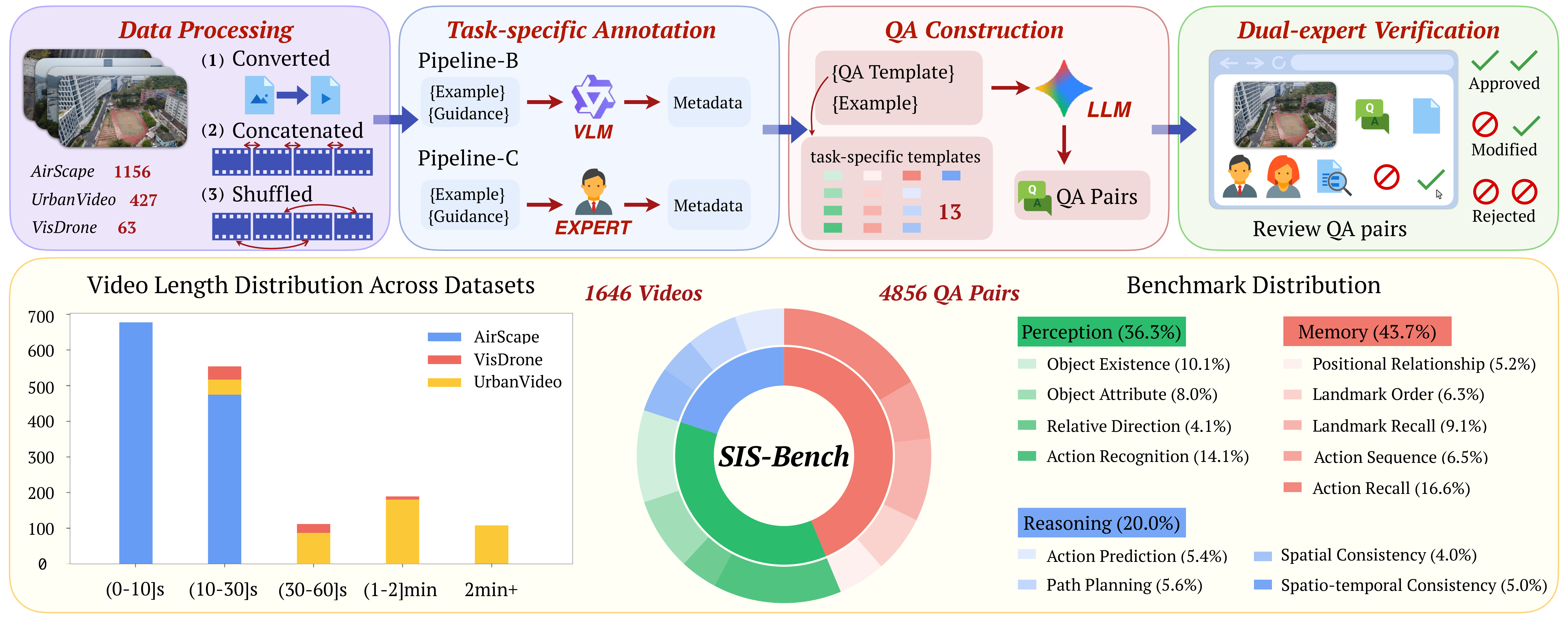}
		\caption{Overview of the SIS-Bench construction pipeline and statistics. The upper panel illustrates the four-stage protocol: Data Processing, Task-specific Annotation, QA Construction, and Dual-expert Verification. The lower panel details dataset statistics, including video-source composition, duration, and task-wise distribution across 13 categories.}
	\label{Fig:pipeline}
    \vspace{-10pt}
\end{figure*}

\subsection{Benchmark Construction Pipeline}
\label{sec:pipeline}

As illustrated in the upper panel of Figure~\ref{Fig:pipeline}, SIS-Bench is constructed with a four-stage pipeline: \textbf{Data Processing}, \textbf{Task-specific Annotation}, \textbf{QA Construction}, and \textbf{Dual-expert Verification}. This protocol is designed to keep the benchmark both scalable and reliable across heterogeneous tasks and datasets.

\noindent \textbf{Data Processing.}
We start from three public sources: AirScape~\cite{zhao2025airscape}, UrbanVideo-Bench~\cite{UrbanVideo-Bench}, and VisDrone~\cite{wen2019visdrone}. As shown in Figure~\ref{Fig:pipeline}, this stage contains three task-oriented operations: \textbf{(1) Conversion}, where VisDrone frame sequences are converted into videos (15 FPS); \textbf{(2) Concatenation}, where 2--4 AirScape clips are stitched to build multi-segment videos for memory evaluation; and \textbf{(3) Shuffling}, where selected long UrbanVideo-Bench videos are reordered through task-specific shuffling strategies to construct spatial reasoning samples. These operations transform heterogeneous source data into capability-aligned video inputs for the later stages.

\noindent \textbf{Task-specific Annotation.}
Following the processed video types, we employ three annotation pipelines, denoted as \textbf{Pipeline-A}, \textbf{Pipeline-B}, and \textbf{Pipeline-C} in Figure~\ref{Fig:pipeline}. \textbf{Pipeline-A} is used for self-awareness perception and memory tasks built from \emph{Single Video} and \emph{Concatenated Video} inputs. Since these samples already contain reliable action annotations, we directly reorganize the existing labels into structured metadata such as action category, action order, and clip-level action correspondence. \textbf{Pipeline-B} is used for spatial perception and memory tasks. We first prompt a VLM to annotate each sample using three types of input: the video itself, task-specific guidance, and in-context examples. The guidance instructs the model to focus on static landmarks and scene elements, maintain unique semantic references to target objects, and summarize the visible entities and their relations in a structured way. The examples demonstrate the expected annotation format and level of detail, so that the model outputs metadata that can be reliably converted into QA pairs; these metadata are then manually filtered and corrected. \textbf{Pipeline-C} is used for reasoning tasks on \emph{Long Video} and \emph{Shuffled Video} inputs, where automatic labeling is more prone to hallucination. For this stage, we develop a multi-annotator collaborative platform, and UAV-experienced experts annotate the samples by following task-specific instructions and examples. The platform further standardizes and organizes the resulting annotations into metadata fields, supporting later QA construction.

\noindent \textbf{QA Construction.}
After annotation, we convert verified metadata into multiple-choice QA pairs through task-specific templates with LLM assistance. The templates are designed to preserve consistent answer formats and difficulty within each task, while distractors are manually controlled to remain plausible but non-ambiguous.

\noindent \textbf{Dual-expert Verification.}
All candidate QA pairs are independently reviewed by two experts on our internal verification platform. Samples are retained only when both reviewers accept them; samples rejected by both are removed; samples with disagreement enter discussion and are either revised and accepted or discarded. This final stage enforces answer uniqueness, question validity, and difficulty consistency across all 13 tasks, and directly supports the evaluation reported in Sec.~\ref{sec:eva}.

\subsection{Benchmark Statistics}
\label{sec:stats}

\noindent \textbf{Multi-level and multi-capability evaluation.}
SIS-Bench evaluates embodied UAV intelligence over two dimensions, \textbf{spatial cognition} and \textbf{self-awareness}, and three cognitive levels, \textbf{perception}, \textbf{memory}, and \textbf{reasoning}. This yields \textbf{13} tasks that jointly cover direct recognition, cross-segment memory, and higher-level inference. As shown in Figure~\ref{Fig:pipeline}, Perception accounts for 36.3\% of SIS-Bench, Memory accounts for 43.7\%, and Reasoning accounts for 20.0\%, yielding a balanced, cognitively progressive benchmark.

\noindent \textbf{Multi-source and multi-type video data.}
SIS-Bench contains \textbf{4,856} multiple-choice question--answer pairs from \textbf{1,646} real-world UAV videos collected from AirScape~\cite{zhao2025airscape}, UrbanVideo-Bench~\cite{UrbanVideo-Bench}, and VisDrone~\cite{wen2019visdrone}. In total, these videos cover approximately \textbf{14.9} hours of UAV footage and span diverse environments, including urban, residential, industrial, and natural scenes. At the video level, the benchmark combines \emph{Single Video}, \emph{Concatenated Video}, \emph{Long Video}, and \emph{Shuffled Video}, covering distinct temporal structures for different tasks. All QA pairs are built through task-matched pipelines and dual-expert verification.

\section{How Well Do MLLMs Handle Self-Awareness and Spatial Cognition in UAV Scenarios?}
\label{sec:eva}

\subsection{Evaluation Setup}

\textbf{Benchmark Models.} We conduct the evaluations on 26 video-capable multimodal large language models (MLLMs), covering a diverse range of model families, parameter scales, and training paradigms. For proprietary models, we evaluate Gemini 3 Flash Preview~\cite{gemini3flash}, GPT-5.4~\cite{gpt5}, Kimi-2.5~\cite{team2026kimi}, Doubao-Seed-1.8~\cite{doubaoseed18}, Doubao-Seed-1.6-Vision~\cite{doubaoseed16vision}, and Qwen3.5-Plus~\cite{qwen35plus}.
For open-source models, we include representative models such as Qwen3-VL~\cite{bai2025qwen3}, InternVL series (2.5–3.5)~\cite{chen2024expanding,zhu2025internvl3,wang2025internvl3}, Kimi-VL~\cite{team2025kimi}, MiMo-VL~\cite{yue2025mimo}, GLM-4V series~\cite{hong2025glm}, Ovis2.5~\cite{lu2025ovis2}, and Step3-VL~\cite{huang2026step3}.

All evaluations are conducted under a zero-shot setting using default prompts provided by each model.
To ensure reproducibility, we adopt greedy decoding for all models.

\noindent\textbf{Evaluation Metric.}
Since all questions are in multiple-choice format, we report accuracy for each task as well as overall accuracy.

\begin{table*}[t]
\centering
\caption{Performance on SIS-Bench. Accuracy (\%) across 13 tasks on 6 proprietary models and 20 open-source models.}
\label{tab:main_results}
\scriptsize
\setlength{\tabcolsep}{2.2pt}
\renewcommand{\arraystretch}{1.05}
\resizebox{\textwidth}{!}{
\begin{tabular}{l|ccc|ccc|ccc|cc|c|cc|cc}
\hline
& & & & \multicolumn{8}{c|}{\textcolor[HTML]{675BC6}{\textbf{Spatial Cognition}}} & \multicolumn{5}{c}{\textcolor[HTML]{FF7F00}{\textbf{Self-Awareness}}} \\
& & & & \multicolumn{3}{c|}{\cellcolor[HTML]{EBF4E2}Perception} & \multicolumn{3}{c|}{\cellcolor[HTML]{FDF0EF}Memory} & \multicolumn{2}{c|}{\cellcolor[HTML]{E5EEFD}Reasoning} & \cellcolor[HTML]{EBF4E2}Perc. & \multicolumn{2}{c|}{\cellcolor[HTML]{FDF0EF}Memory} & \multicolumn{2}{c}{\cellcolor[HTML]{E5EEFD}Reasoning} \\
\textbf{Model} & \rotatebox{90}{\textbf{Perc.}} & \rotatebox{90}{\textbf{Perc.+Mem.}} & \rotatebox{90}{\textbf{Overall}} & \rotatebox{90}{\textit{Obj Exist}} & \rotatebox{90}{\textit{Obj Attr}} & \rotatebox{90}{\textit{Rel Dir}} & \rotatebox{90}{\textit{Land Order}} & \rotatebox{90}{\textit{Land Recall}} & \rotatebox{90}{\textit{Pos Rel}} & \rotatebox{90}{\textit{Spat Consist}} & \rotatebox{90}{\textit{ST Consist}} & \rotatebox{90}{\textit{Act Recog}} & \rotatebox{90}{\textit{Act Seq}} & \rotatebox{90}{\textit{Act Recall}} & \rotatebox{90}{\textit{Act Pred}} & \rotatebox{90}{\textit{Path Plan}} \\
\hline
Random & 24.8 & 25.0 & 25.0 & 24.7 & 24.9 & 24.9 & 25.2 & 25.2 & 25.1 & 25.0 & 25.2 & 24.7 & 25.3 & 25.0 & 24.9 & 25.2 \\
Human Performance & 93.4 & 94.2 & 91.7 & 95.9 & 96.6 & 84.4 & 97.0 & 93.8 & 94.6 & 88.3 & 68.8 & 92.4 & 96.6 & 94.1 & 86.9 & 83.6 \\
\hline
\multicolumn{17}{l}{\cellcolor[HTML]{ECF4FF}\textit{Proprietary Models}} \\
Gemini-3-Flash & 79.1 & 74.4 & 71.6 & 97.2 & 80.6 & 75.0 & 94.4 & 84.0 & 89.7 & 71.8 & 57.7 & 66.5 & 84.1 & 42.4 & 61.6 & 53.7 \\
Kimi-2.5 & 73.0 & 73.2 & 71.0 & 97.0 & 78.0 & 80.5 & 98.0 & 81.9 & 87.3 & 76.4 & 53.1 & 50.9 & 77.8 & 53.0 & 65.4 & 57.0 \\
Doubao-Seed-1.8 & 68.7 & 73.3 & 70.6 & 97.4 & 72.6 & 76.5 & 96.1 & 84.4 & 87.7 & 66.2 & 57.3 & 43.7 & 85.1 & 59.2 & 63.1 & 54.0 \\
Qwen3.5-Plus & 73.3 & 71.7 & 70.1 & 97.6 & 76.2 & 78.5 & 97.7 & 83.3 & 87.7 & 75.9 & 54.8 & 52.8 & 82.5 & 42.4 & 69.2 & 58.5 \\
GPT-5.4 & 73.7 & 72.2 & 70.0 & 98.4 & 74.9 & 67.0 & 94.8 & 78.6 & 86.9 & 64.1 & 57.3 & 57.3 & 77.8 & 49.9 & 65.8 & 58.8 \\
Doubao-Seed-1.6-Vision & 72.8 & 71.8 & 69.8 & 97.4 & 73.9 & 78.0 & 94.4 & 81.0 & 87.7 & 72.3 & 53.5 & 53.1 & 77.1 & 48.9 & 65.8 & 57.4 \\

\hline
\multicolumn{17}{l}{\cellcolor[HTML]{ECF4FF}\textit{Open-source Models}} \\
Qwen3-VL-8B-Instruct & 74.9 & 67.0 & 64.6 & 97.2 & 82.2 & 73.5 & 95.8 & 74.9 & 82.9 & 64.6 & 51.0 & 55.1 & 60.3 & 32.2 & 57.0 & 48.9 \\
Qwen3-VL-8B-Thinking & 72.3 & 66.5 & 64.7 & 95.5 & 76.0 & 72.5 & 95.1 & 78.6 & 79.8 & 64.6 & 55.6 & 53.4 & 62.9 & 33.6 & 64.6 & 46.3 \\
Qwen3-VL-4B-Instruct & 73.3 & 65.1 & 62.5 & 96.3 & 80.4 & 71.0 & 93.8 & 70.7 & 82.1 & 73.3 & 47.3 & 53.4 & 61.3 & 29.1 & 57.0 & 36.4 \\
InternVL3.5-8B & 73.4 & 64.0 & 61.1 & 96.1 & 76.5 & 70.5 & 86.3 & 72.0 & 73.8 & 55.4 & 47.3 & 56.1 & 52.4 & 31.8 & 55.1 & 41.9 \\
Qwen3-VL-30B-A3B-Instruct & 68.7 & 62.5 & 61.0 & 97.2 & 78.6 & 79.0 & 95.1 & 76.5 & 85.3 & 67.7 & 47.7 & 39.7 & 44.8 & 28.4 & 59.7 & 48.9 \\
GLM-4.1V-9B-Thinking & 69.5 & 62.9 & 60.4 & 92.1 & 77.5 & 72.0 & 93.5 & 64.1 & 75.4 & 63.6 & 49.8 & 48.1 & 55.9 & 34.8 & 48.7 & 43.4 \\
Step3-VL-10B & 69.6 & 62.1 & 60.3 & 95.1 & 72.4 & 68.5 & 94.1 & 72.2 & 67.5 & 64.1 & 54.4 & 50.0 & 55.9 & 28.6 & 54.4 & 42.3 \\
MiMo-VL-7B-RL & 68.9 & 61.1 & 59.2 & 96.3 & 74.2 & 73.5 & 80.1 & 75.2 & 81.3 & 72.3 & 47.3 & 44.8 & 43.2 & 29.7 & 51.7 & 40.8 \\
GLM-4.6V-Flash-9B & 73.2 & 61.9 & 59.6 & 96.5 & 79.1 & 66.0 & 89.5 & 44.9 & 65.1 & 64.6 & 41.9 & 55.2 & 56.2 & 37.1 & 55.1 & 43.4 \\
InternVL3-14B & 68.5 & 61.6 & 59.1 & 96.3 & 77.5 & 71.0 & 94.8 & 79.7 & 81.3 & 67.7 & 46.1 & 42.7 & 43.8 & 24.9 & 55.5 & 31.6 \\
Kimi-VL-A3B-Instruct & 68.5 & 61.7 & 59.0 & 95.1 & 76.5 & 70.0 & 87.9 & 74.3 & 79.4 & 66.7 & 45.6 & 44.6 & 42.5 & 31.8 & 57.4 & 27.9 \\
MiMo-VL-7B-SFT & 68.7 & 60.7 & 58.8 & 96.3 & 73.9 & 72.0 & 80.7 & 72.0 & 80.6 & 71.3 & 44.4 & 44.9 & 43.8 & 29.7 & 51.3 & 42.3 \\
Ovis2.5-9B & 71.8 & 60.3 & 58.5 & 96.7 & 79.1 & 69.5 & 76.5 & 63.9 & 73.0 & 58.5 & 46.1 & 50.6 & 50.5 & 26.7 & 60.1 & 41.5 \\
InternVL2.5-8B & 71.0 & 61.0 & 58.2 & 95.3 & 82.9 & 67.0 & 85.3 & 66.6 & 79.0 & 61.0 & 50.2 & 48.0 & 29.5 & 33.2 & 53.2 & 29.0 \\
InternVL3.5-4B & 72.3 & 60.4 & 58.0 & 96.5 & 78.3 & 70.5 & 88.9 & 61.2 & 71.0 & 55.4 & 51.9 & 52.0 & 33.3 & 30.2 & 56.7 & 33.1 \\
InternVL3-9B & 67.2 & 59.3 & 57.3 & 95.3 & 73.6 & 74.0 & 84.3 & 71.1 & 79.0 & 63.1 & 46.5 & 41.4 & 38.4 & 28.2 & 57.0 & 34.6 \\
Qwen2.5-VL-7B-Instruct & 67.1 & 57.5 & 55.8 & 96.5 & 71.1 & 69.0 & 73.2 & 54.2 & 69.0 & 67.7 & 41.5 & 43.3 & 55.6 & 29.2 & 60.8 & 31.6 \\
Qwen2.5-VL-3B-Instruct & 58.6 & 56.2 & 53.6 & 92.7 & 48.8 & 64.5 & 90.8 & 51.9 & 62.7 & 57.4 & 42.7 & 37.9 & 62.5 & 35.6 & 53.6 & 23.2 \\
Qwen2-VL-7B-Instruct & 59.5 & 54.8 & 53.2 & 96.7 & 48.3 & 61.0 & 71.6 & 58.0 & 69.0 & 63.6 & 47.3 & 38.6 & 51.4 & 33.3 & 58.9 & 22.1 \\
Qwen3-VL-2B-Instruct & 63.0 & 54.2 & 50.5 & 97.4 & 76.2 & 63.0 & 54.2 & 53.3 & 74.2 & 61.5 & 26.1 & 31.0 & 54.0 & 29.2 & 42.2 & 19.9 \\

\hline
\end{tabular}
}
\end{table*}

\subsection{Evaluation on SIS-Bench}

Table~\ref{tab:main_results} summarizes the performance of all evaluated models on SIS-Bench.
Overall, the results reveal a consistent pattern across model families: current MLLMs are substantially stronger at modeling the external environment than the embodied agent itself, and their performance degrades progressively as tasks require longer temporal integration and higher-order reasoning.

\noindent\textbf{Overall performance and human upper bound.}
Human evaluators achieve an overall accuracy of \textbf{91.7\%}, substantially outperforming all evaluated models; the best model reaches \textbf{71.6\%}, leaving a gap of over \textbf{20} points. This shows SIS-Bench is far from saturated, with a large gap to human embodied spatial understanding. The gap is especially evident on self-awareness and reasoning tasks, indicating that current models still struggle to form coherent, temporally grounded representations of the UAV as an embodied agent.

\noindent\textbf{Imbalance between spatial cognition and self-awareness.}
A key finding of SIS-Bench is the imbalance between \emph{spatial cognition} and \emph{self-awareness}. Across most evaluated models, performance on spatial cognition tasks is consistently higher than on self-awareness tasks. In other words, current MLLMs are better at interpreting the external scene---such as object layouts, attributes, and landmark relations---than at modeling the UAV's own state, motion history, and action dynamics. This directly supports our motivation in Sec.~\ref{sec:intro}: existing UAV MLLMs focus heavily on environmental perception, often at the expense of embodied-agent awareness.

\noindent\textbf{Progressive degradation across the cognitive hierarchy.}
The results also show a clear cognitive hierarchy: performance is highest on \emph{perception}, drops on \emph{memory}, and declines further on \emph{reasoning}. This trend is consistent across both proprietary and open-source models. Perception mainly depends on direct visual recognition, memory requires retaining temporally distributed evidence, and reasoning requires integrating such evidence into coherent spatial or action-level inference. This degradation suggests current MLLMs are much better at immediate recognition than at maintaining stable representations over time and reasoning over them.

\noindent\textbf{Model families and the limitation of scaling.}
Proprietary models generally outperform open-source models on overall accuracy, showing benefits from stronger pretraining and system-level optimization. However, this gap is still modest compared with the large remaining gap to human performance. More importantly, gains from stronger models are uneven across tasks: clearer on reasoning-intensive spatial cognition tasks, but much less evident on UAV motion, action recognition, and action recall. This suggests scaling improves general recognition and inference, but does not fundamentally resolve dynamic, agent-centered modeling.

\noindent\textbf{Implications for embodied UAV intelligence.}
Taken together, the above results suggest that current MLLMs still lack explicit mechanisms for modeling self-related dynamics in UAV scenarios. Changes in the UAV's self-state directly alter how the external environment is observed; for example, when the UAV moves rapidly forward, surrounding scene elements appear to move backward in the visual field. This naturally raises the next question: can stronger motion-aware cues improve the joint modeling of \emph{self} and \emph{space}? We investigate this question in the following section through a controlled motion-aware exploration.

\section{A Motion-aware Exploration of Self in Space}
\label{sec:method}

\begin{figure*}[t]
	\centering
	\includegraphics[width = 0.95\linewidth]{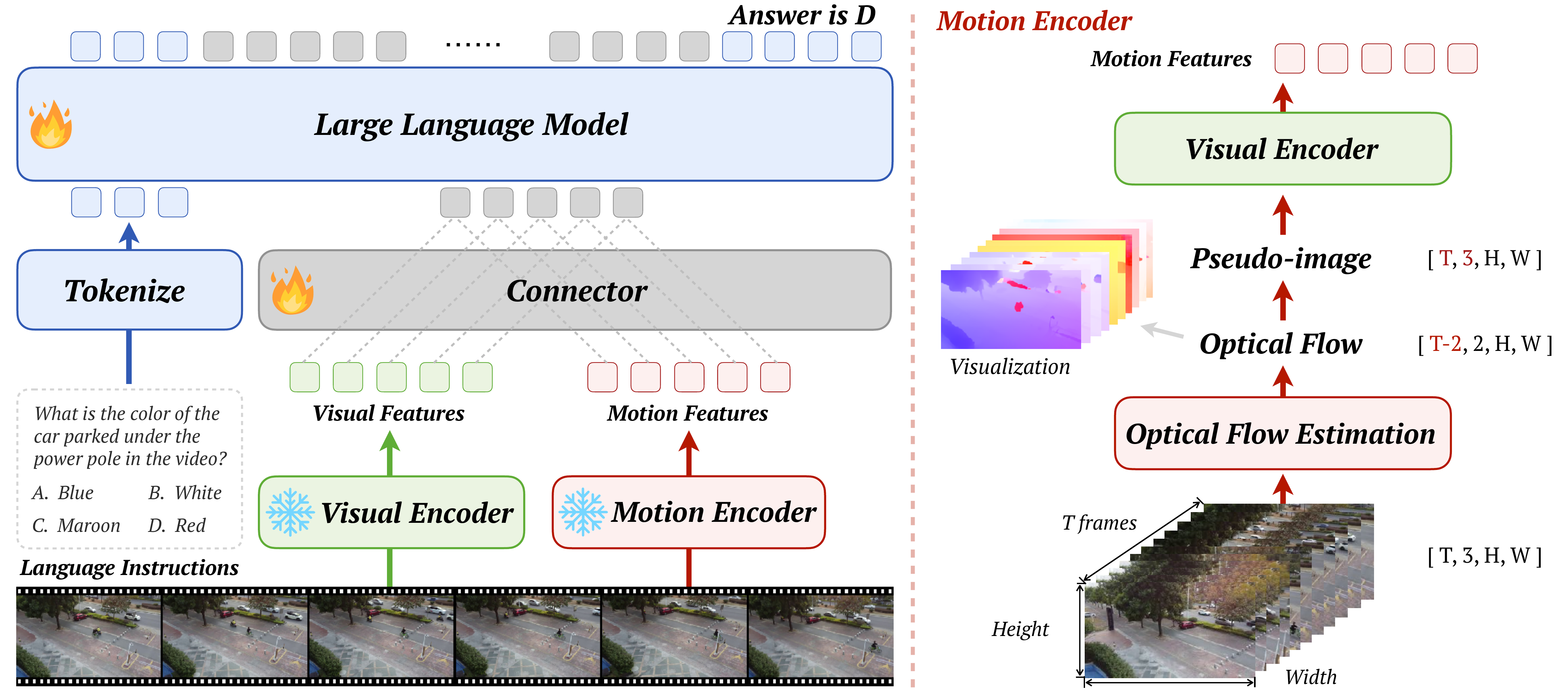}
	\caption{Overview of the SIS-Motion framework. The architecture (left) comprises parallel encoders for synchronized visual and motion processing. The internal mechanism of the Motion Encoder is further elaborated on the right, showing the structural alignment of motion features into unified tokens compatible with large-scale language models.}
    
	\label{Fig:framework}
    \vspace{-10pt}
\end{figure*}

Spatial cognition and self-awareness should be inherently coupled in embodied UAV intelligence. In real aerial operation, the UAV's self-state is always situated in space: viewpoint changes, scene evolution, and agent motion are tightly intertwined with the surrounding environment. This raises a natural question: \emph{can explicitly modeling motion-related changes in space help MLLMs better understand self, and can such self-related modeling also benefit spatial cognition?}

To investigate this question, we conduct a motion-aware exploration. Specifically, we instantiate \textbf{SIS-Motion} as a motion-aware extension of a standard video MLLM by augmenting it with an optical-flow-based motion encoder and a lightweight fusion connector, so that self-related motion cues can be integrated with appearance-based spatial context. As shown below, this exploration yields two main findings: (1) integrating motion cues improves both self-awareness and spatial cognition on SIS-Bench; and (2) the gains are most evident on perception- and memory-level tasks, while also transferring to downstream UAV navigation.

\subsection{SIS-Motion Framework}

We implement SIS-Motion as a motion-aware extension of a standard video MLLM. The key idea is to complement appearance-based visual representations with explicit motion cues, so that the model can jointly capture scene semantics and self-related dynamics under changing viewpoints. The architecture of SIS-Motion is shown in Figure~\ref{Fig:framework}, where we adopt a dual-encoder design to jointly model environmental context and self-motion, consisting of a visual encoder, a motion encoder, and a connector. 

\noindent \textbf{Visual Encoder.}
Given an input video clip \( \{\mathbf{f}_i\}_{i=1}^{N} \), where \( \mathbf{f}_i\in\mathbb{R}^{3\times H\times W} \), the visual encoder \( \mathcal{E}_{v} \) extracts semantic video tokens:
\[
\mathbf{e}_{v}=\mathcal{E}_{v}(\{\mathbf{f}_i\}_{i=1}^{N}), \quad
\mathbf{e}_{v}\in\mathbb{R}^{N_v\times d}.
\]
These tokens mainly encode scene appearance, object layout, and environmental context.

\noindent\textbf{Motion Encoder.}
To explicitly model temporal dynamics, we estimate optical flow from the input video:
\[
\mathbf{U}=\Phi(\{\mathbf{f}_i\}_{i=1}^{N}), \quad
\mathbf{U}\in\mathbb{R}^{T_f\times 2\times H_f\times W_f},
\]
where \( \Phi \) denotes an optical-flow estimator (e.g., RAFT~\cite{teed2020raft}). We then convert the flow into a 3-channel pseudo-image sequence:
\[
\mathbf{P}_t=\left[\left\|\mathbf{U}_t\right\|_2,\ \mathbf{U}_t^{x},\ \mathbf{U}_t^{y}\right], \quad
\mathbf{P}\in\mathbb{R}^{T_f\times 3\times H_f\times W_f}.
\]
After temporal alignment to the video length \(N\), the sequence $\mathbf{P}$ is fed into a ViT-based motion encoder \( \mathcal{E}_m \) to obtain motion tokens:
\[
\mathbf{e}_m=\mathcal{E}_m(\mathbf{P}), \quad
\mathbf{e}_m\in\mathbb{R}^{N_v\times d}.
\]
Compared with appearance features, these motion tokens provide more explicit cues about directional movement, action transitions, and viewpoint evolution.

\noindent\textbf{Feature fusion.}
We integrate the two streams through a lightweight connector. Specifically, motion features are first aligned to the visual token space and then fused with visual tokens:
\[
\mathbf{e}_{m}'=\mathrm{MLP}_{m}(\mathrm{LN}(\mathbf{e}_{m})), \quad
\mathbf{e}=\mathbf{e}_{v}+\mathbf{e}_{m}'.
\]
The fused representation \( \mathbf{e} \) is finally fed into the language model for downstream prediction. This design keeps the setup controlled, so that any observed gain can be more directly attributed to motion-aware cues in the joint modeling of \emph{self} and \emph{space}.

\begin{figure}[t]
	\centering
	\includegraphics[width = 0.5\linewidth]{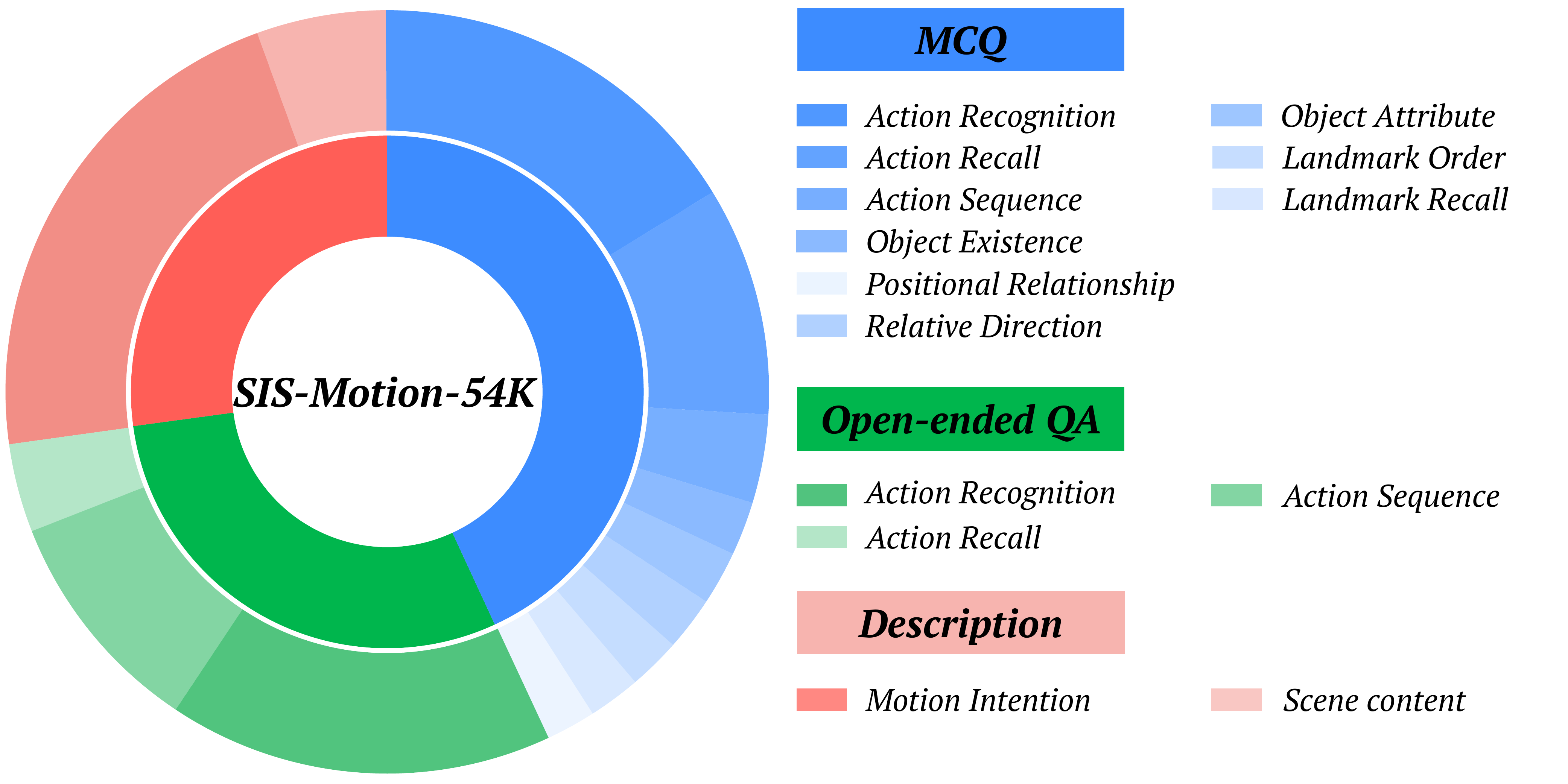}
		\caption{Task distribution of SIS-Motion-54K across three formats: MCQ, Open-ended QA, and Description.}
	\label{Fig:data}
	\vspace{-20pt}
\end{figure}

\subsection{SIS-Motion-54K and Controlled Evaluation}

\noindent \textbf{Training dataset SIS-Motion-54K.}
To support motion-aware training, we construct \textbf{SIS-Motion-54K} from the AirScape~\cite{zhao2025airscape} \textbf{training split}, strictly separated from SIS-Bench. As shown in Figure~\ref{Fig:data}, the dataset contains 54k samples in three complementary formats. \textit{(1) Multiple-choice Question Answering~(MCQ):} covering 9 perception and memory tasks (e.g., Object Existence, Action Recognition), this format aligns training with the discriminative evaluation setting of SIS-Bench, directly strengthening motion-grounded perception. \textit{(2) Open-ended QA:} free-form supervision derived from action-related tasks, encouraging the model to explicitly express motion-aware content. \textit{(3) Description:} \textit{Motion Intention} and \textit{Scene Content} descriptions that provide continuous supervision for jointly grounding agent dynamics and scene semantics. 
Collectively, these three formats provide complementary supervision for motion and spatial grounding. Notably, because the training set focuses strictly on perception and memory rather than high-level reasoning, any subsequent performance gains on reasoning tasks can be robustly attributed to improved spatial-temporal generalization rather than task-specific memorization.

\noindent \textbf{Experiment setup.}
We adopt Qwen2.5-VL 3B~\cite{bai2025qwen2} as the backbone and compare two settings under identical training configurations: a \textbf{visual-only baseline} and a \textbf{motion-aware variant}. In both settings, the visual encoder is frozen and only lightweight adaptation modules are trained, ensuring that the comparison isolates the effect of motion-aware integration rather than large-scale retraining.

\noindent \textbf{Testing protocol.}
We evaluate the resulting models from two perspectives. First, we test them on \textbf{SIS-Bench} to examine whether motion-aware integration improves the joint modeling of \emph{self} and \emph{space}, especially across different cognitive levels. Second, we test transferability on a \textbf{downstream UAV navigation} benchmark constructed from OpenUAV~\cite{wang2024towards}, where continuous path trajectories are converted into multiple-choice action decisions. Together, these two evaluations allow us to examine both benchmark-level gains and their transfer to practical UAV decision-making.

\setlength{\tabcolsep}{3pt}
\begin{table*}[t]
\centering
\caption{Results of the motion-aware exploration on SIS-Bench. Accuracy (\%) across 13 tasks, comparing the zero-shot baseline, visual-only SFT, and SIS-Motion.}
\label{tab:ablation}
\scriptsize
\renewcommand{\arraystretch}{1.05}
\resizebox{\textwidth}{!}{
\begin{tabular}{l|ccc|ccc|ccc|cc|c|cc|cc}
\hline
& & & & \multicolumn{8}{c|}{\textcolor[HTML]{675BC6}{\textbf{Spatial Cognition}}} & \multicolumn{5}{c}{\textcolor[HTML]{FF7F00}{\textbf{Self-Awareness}}} \\
& & & & \multicolumn{3}{c|}{\cellcolor[HTML]{EBF4E2}Perception} & \multicolumn{3}{c|}{\cellcolor[HTML]{FDF0EF}Memory} & \multicolumn{2}{c|}{\cellcolor[HTML]{E5EEFD}Reasoning} & \cellcolor[HTML]{EBF4E2}Perc. & \multicolumn{2}{c|}{\cellcolor[HTML]{FDF0EF}Memory} & \multicolumn{2}{c}{\cellcolor[HTML]{E5EEFD}Reasoning} \\
\textbf{Model} & \rotatebox{90}{\textbf{Spatial Avg}} & \rotatebox{90}{\textbf{Self Avg}} & \rotatebox{90}{\textbf{Overall}} & \rotatebox{90}{\textit{Obj Exist}} & \rotatebox{90}{\textit{Obj Attr}} & \rotatebox{90}{\textit{Rel Dir}} & \rotatebox{90}{\textit{Land Order}} & \rotatebox{90}{\textit{Land Recall}} & \rotatebox{90}{\textit{Pos Rel}} & \rotatebox{90}{\textit{Spat Consist}} & \rotatebox{90}{\textit{ST Consist}} & \rotatebox{90}{\textit{Act Recog}} & \rotatebox{90}{\textit{Act Seq}} & \rotatebox{90}{\textit{Act Recall}} & \rotatebox{90}{\textit{Act Pred}} & \rotatebox{90}{\textit{Path Plan}} \\
\hline

\multicolumn{17}{l}{\cellcolor[HTML]{ECF4FF}\textit{Proprietary Models}} \\
Gemini-3-Flash & 83.7 & 58.6 & 71.6 & 97.2 & 80.6 & 75.0 & 94.4 & 84.0 & 89.7 & 71.8 & 57.7 & 66.5 & 84.1 & 42.4 & 61.6 & 53.7 \\

Doubao-Seed-1.8 & 82.2 & 58.0 & 70.6 & 97.4 & 72.6 & 76.5 & 96.1 & 84.4 & 87.7 & 66.2 & 57.3 & 43.7 & 85.1 & 59.2 & 63.1 & 54.0 \\
Qwen3.5-Plus & 83.5 & 55.7 & 70.1 & 97.6 & 76.2 & 78.5 & 97.7 & 83.3 & 87.7 & 75.9 & 54.8 & 52.8 & 82.5 & 42.4 & 69.2 & 58.5 \\
GPT-5.4 & 80.6 & 58.6 & 70.0 & 98.4 & 74.9 & 67.0 & 94.8 & 78.6 & 86.9 & 64.1 & 57.3 & 57.3 & 77.8 & 49.9 & 65.8 & 58.8 \\

\hline

\multicolumn{17}{l}{\cellcolor[HTML]{ECF4FF}\textit{Baseline}} \\
ZeroShot & 65.8 & 40.5 & 53.6 & 92.7 & 48.8 & 64.5 & 90.8 & 51.9 & 62.7 & 57.4 & 42.7 & 37.9 & 62.5 & 35.6 & 53.6 & 23.2 \\
SFT & 72.0 & 60.3 & 66.4 & 96.1 & 67.7 & 66.0 & 93.8 & 63.4 & 69.8 & 57.4 & 36.9 & 88.0 & 80.6 & 47.1 & 45.2 & 20.6 \\
\hline

\multicolumn{17}{l}{\cellcolor[HTML]{ECF4FF}\textit{Proprietary (Ours)}} \\
SIS-Motion & 74.2 & 63.7 & 69.1 & 96.5 & 71.1 & 66.0 & 93.8 & 68.2 & 73.0 & 60.0 & 39.8 & 88.1 & 81.3 & 55.9 & 44.1 & 23.5 \\
\hline
\end{tabular}
}
\end{table*}

\subsection{Motion-aware Results}

Table~\ref{tab:ablation} reports the results of our motion-aware exploration on SIS-Bench, where we derive two main findings.

\noindent \textbf{Coupling self and space improves both.}
Compared with the visual-only SFT baseline, SIS-Motion improves both Spatial Avg and Self Avg, from 72.0 to 74.2 and from 60.3 to 63.7, respectively. These gains indicate that explicitly modeling self-related motion benefits not only understanding of the UAV itself, but also interpretation of the surrounding environment under changing viewpoints. In other words, the effect of motion-aware modeling is shared by both \emph{self} and \emph{space}, rather than being confined to one side.

\noindent \textbf{Perception and memory benefits more than reasoning.}
The improvements are most evident on perception- and memory-related tasks, including Object Attribute, Landmark Recall, Positional Relationship, Action Recognition, Action Sequence, and Action Recall. By contrast, the gains on reasoning tasks are limited and less consistent: SIS-Motion improves Path Planning, but does not improve Action Prediction. This pattern is consistent with both the design of SIS-Motion-54K and the diagnosis in Sec.~\ref{sec:eva}. Since the supplementary dataset mainly strengthens motion-grounded perception, temporal continuity, and action-related understanding, it naturally benefits perception and memory first, while higher-level reasoning remains more challenging.

\noindent \textbf{Robust across different motion encoders.} We further test whether the above findings are robust across different motion encoders by replacing the optical-flow estimator~(RAFT~\cite{teed2020raft}, Sea-RAFT~\cite{wang2024sea}, MemFlow~\cite{dong2024memflow}, and MOFNet~\cite{shi2023videoflow}) while keeping the rest of the setup fixed. All four variants outperform the visual-only SFT baseline (66.4\% overall in Table~\ref{tab:ablation}), indicating that the gains brought by motion-aware integration are robust to the choice of motion encoder. Better motion quality further improves performance, with MOFNet achieving the best result.

\setlength{\tabcolsep}{3pt}
\begin{table}[t]
\centering
\caption{Ablation on optical-flow estimators under the same SIS-Motion framework.}
\label{tab:flow_ablation}
\scriptsize
\renewcommand{\arraystretch}{1.05}
\scalebox{1.4}{
\begin{tabular}{l|c|c|c}
\hline
\scalebox{0.75}{\textbf{Flow Estimation}} & \scalebox{0.75}{\textbf{Spatial Avg.}} & \scalebox{0.75}{\textbf{Awareness Avg.}} & \scalebox{0.75}{\textbf{Overall}} \\
\hline
RAFT\cite{teed2020raft} & 73.8 & 59.9 & 67.1 \\
Sea-RAFT\cite{wang2024sea} & 74.6 & 61.0 & 68.1 \\
MemFlow~\cite{dong2024memflow} & 73.9 & 62.2 & 68.2 \\
MOFNet\cite{shi2023videoflow} & 74.2 & 63.7 & 69.1 \\
\hline
\end{tabular}
}
\end{table}

\subsection{Transfer to Downstream UAV Navigation}

\begin{figure}[H]
\centering
\includegraphics[width=\textwidth]{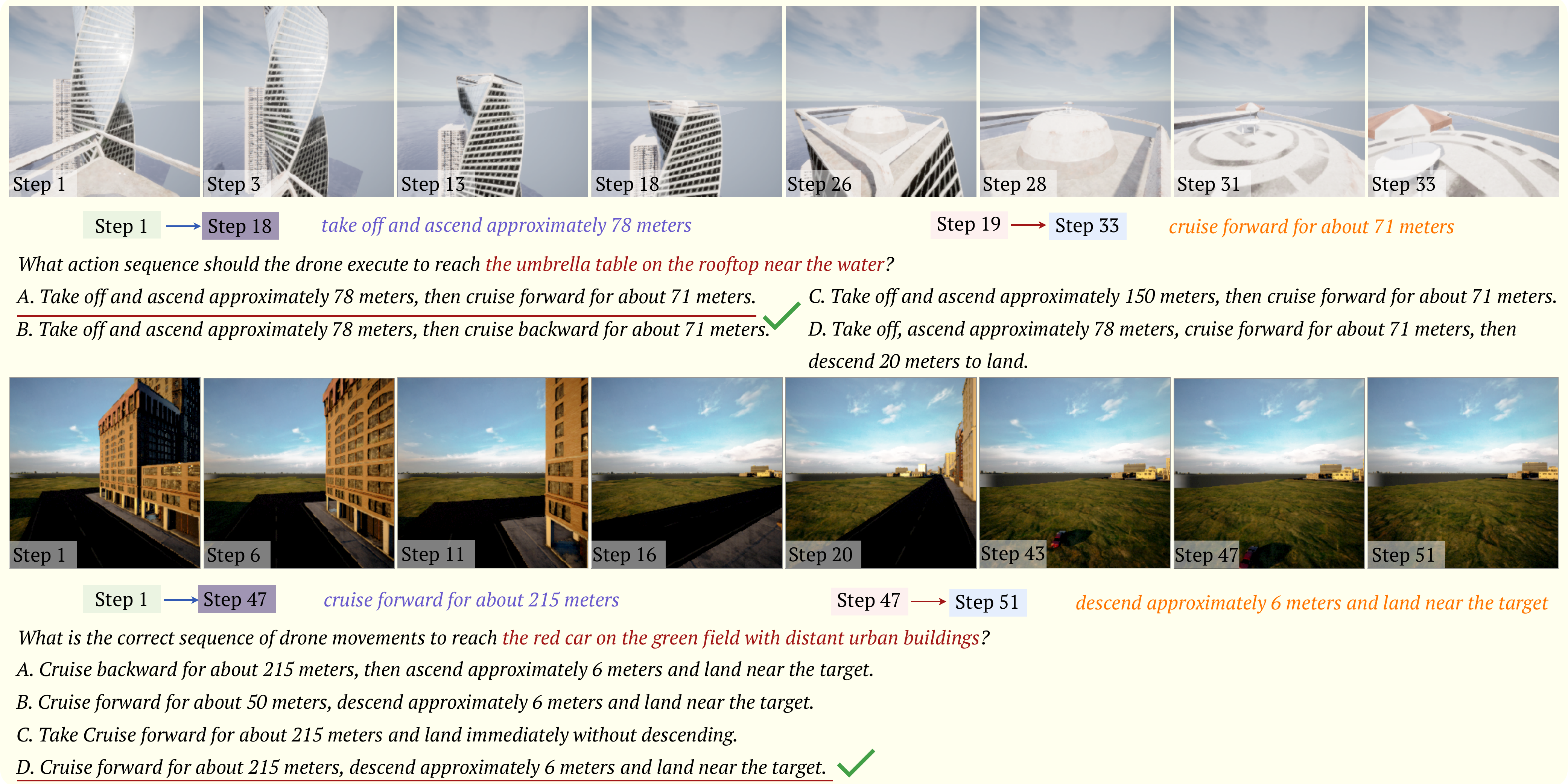}
\Description{An overview of the downstream OpenUAV-based path-planning task, showing front-view trajectory frames, rule-based operation extraction, conversion into multiple-choice questions, and evaluation examples.}
\caption{Scenarios of the downstream UAV navigation task. Representative samples evaluate the drone's decision-making and path-planning capabilities by requiring the model to choose the optimal flight sequence to reach a designated target.}
\label{fig:downstream_task}
\end{figure}

To further examine whether the benefit of motion-aware self-in-space modeling transfers beyond our benchmark, we evaluate on a downstream UAV navigation task built from OpenUAV~\cite{wang2024towards}. Representative navigation scenarios are shown in Figure~\ref{fig:downstream_task}.

The benchmark requires the model to make action decisions under realistic navigation contexts and contains 3{,}895 questions from 22 simulated environments, covering urban, desert, forest, port, and island scenes. All models are evaluated in a zero-shot manner without further fine-tuning, and more details are provided in the supplementary material. The results show that SIS-Motion achieves an accuracy of 92.2\%, significantly outperforming the Qwen2.5-VL 3B backbone (71.2\%). This suggests that the gains from motion-aware modeling are not limited to our SIS-Bench, but can also transfer to practical downstream UAV navigation scenarios.

\section{Conclusion}

In this work, we introduce \textbf{SIS-Bench}, a benchmark for embodied UAV intelligence under a unified \emph{self-in-space} formulation. SIS-Bench is built through a task-conditioned construction pipeline with heterogeneous video processing, task-specific annotation and dual-expert verification. It contains 4,856 question--answer pairs from 1,646 real-world UAV videos spanning 14.9 hours of aerial footage, and organizes evaluation into 13 tasks over two complementary dimensions, \emph{spatial cognition} and \emph{self-awareness}, and three cognitive levels, \emph{perception}, \emph{memory}, and \emph{reasoning}. Based on this benchmark, we find that current MLLMs exhibit two consistent limitations in UAV embodied scenarios: they are substantially stronger at modeling space than self, and their performance degrades progressively from perception to memory to reasoning. Motivated by these findings, we further conduct a controlled motion-aware exploration through \textbf{SIS-Motion}. The results show that self-related motion improves UAV self-understanding and spatial interpretation under changing viewpoints, with transfer to downstream UAV navigation, underscoring the importance of unified \emph{self-in-space} modeling for embodied UAV intelligence.

\clearpage
\bibliographystyle{plainnat}
\bibliography{uavideo}

\clearpage
\beginappendix

\begin{figure}[H]
\centering
\includegraphics[width=\textwidth]{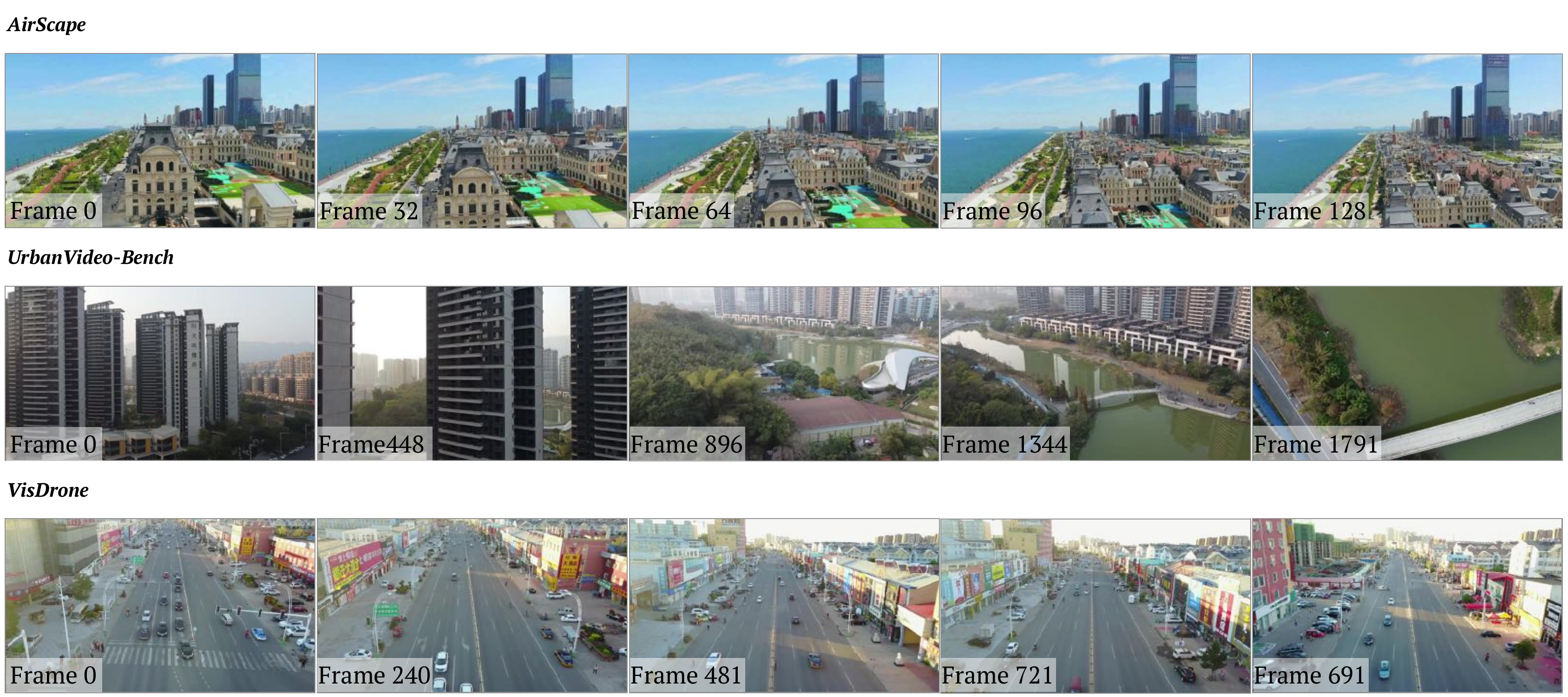}
\Description{Three rows of representative source-video frames from AirScape, UrbanVideo-Bench, and VisDrone. Each row contains five sampled frames with frame indices overlaid to indicate temporal positions in the original video.}
\caption{Representative source videos used in SIS-Bench construction. Each row shows five temporally sampled frames from one example source video, with frame indices overlaid to indicate their original positions. From top to bottom, AirScape provides short action-centric clips, UrbanVideo-Bench provides long real-world urban flights with substantial viewpoint change, and VisDrone provides low-altitude urban street observations collected from frame sequences.}
\label{fig:source_video_examples}
\end{figure}

\section{Benchmark Construction Details}
\label{sec:video_source_construction}

This section supplements the benchmark construction description in the main paper by detailing the source datasets and video processing steps used in benchmark construction. Through these steps, we obtain four types of videos in the final benchmark: \emph{Single Video}, \emph{Concatenated Video}, \emph{Long Video}, and \emph{Shuffled Video}. 

\subsection{Source datasets and collection protocol}

To construct a benchmark that comprehensively evaluates embodied spatial intelligence, we select source datasets based on three complementary criteria: (1) \textbf{temporal characteristics}, including both short clips and long continuous videos to support perception, memory, and reasoning evaluation; (2) \textbf{motion and action diversity}, enabling the modeling of agent dynamics and self-related processes; and (3) \textbf{real-world scene diversity}, prioritizing authentic UAV observations from varied real-world environments to ensure practical relevance and broad generalization. Based on these criteria, we combine AirScape, UrbanVideo-Bench, and VisDrone, which together provide complementary strengths in action annotations, long-term temporal continuity, and diverse urban scenes.

\paragraph{AirScape.}
AirScape is a UAV dataset with action-intent annotations, and its videos are generally short. The dataset is built from three underlying sources: UrbanVideo, NAT2021, and WebUAV3M. We use the original AirScape test split as the main source pool for benchmark construction and reserve the original training split for subsequent training-data construction. However, since we additionally use UrbanVideo-Bench as an independent benchmark source, we repartition AirScape to avoid test leakage from the UrbanVideo subset. The original and adjusted splits are shown in Table~\ref{tab:airscape_split}. In the adjusted protocol, all UrbanVideo-origin clips are placed on the benchmark side, the NAT2021 split remains unchanged, and the WebUAV3M portion is re-balanced so that the benchmark retains 700 clips while the remaining clips are moved to the training pool.

\begin{table}[b]
\centering
\caption{Original and adjusted AirScape splits used in our benchmark construction.}
\label{tab:airscape_split}
\small
\begin{tabular}{lccc}
\hline
\textbf{Split} & \textbf{UrbanVideo} & \textbf{NAT2021} & \textbf{WebUAV3M} \\
\hline
Original test & 140 & 444 & 2538 \\
Original train & 783 & 1349 & 5758 \\
\hline
Adjusted benchmark & 923 & 444 & 700 \\
Adjusted training & 0 & 1349 & 7596 \\
\hline
\end{tabular}
\end{table}

\paragraph{UrbanVideo-Bench.}
UrbanVideo-Bench contains both real-world and simulated UAV data. In this work, we select 174 real-world videos from UrbanVideo-Bench. These videos are recorded in authentic urban environments in Shenzhen, Guangdong Province, China, including city streets, residential communities, schools, and parks. Compared with the short AirScape clips, these real-world UrbanVideo-Bench videos are typically much longer and preserve continuous flight trajectories, making them particularly suitable for \emph{Long Video} and \emph{Shuffled Video} construction.

\paragraph{VisDrone.}
We collect 63 scene-level frame sequences from VisDrone. The content mainly consists of low-altitude UAV views over urban areas, such as streets, parks, and residential neighborhoods. These sequences have moderate temporal length and enrich the scene diversity of the benchmark. Since VisDrone is released as ordered frames rather than ready-to-use videos for our setting, we first convert each sequence into a video before later annotation and QA construction. Representative sampled frames from the three source datasets are shown in Figure~\ref{fig:source_video_examples}.

\subsection{Video construction protocol}

\begin{figure}[H]
\centering
\includegraphics[width=\textwidth]{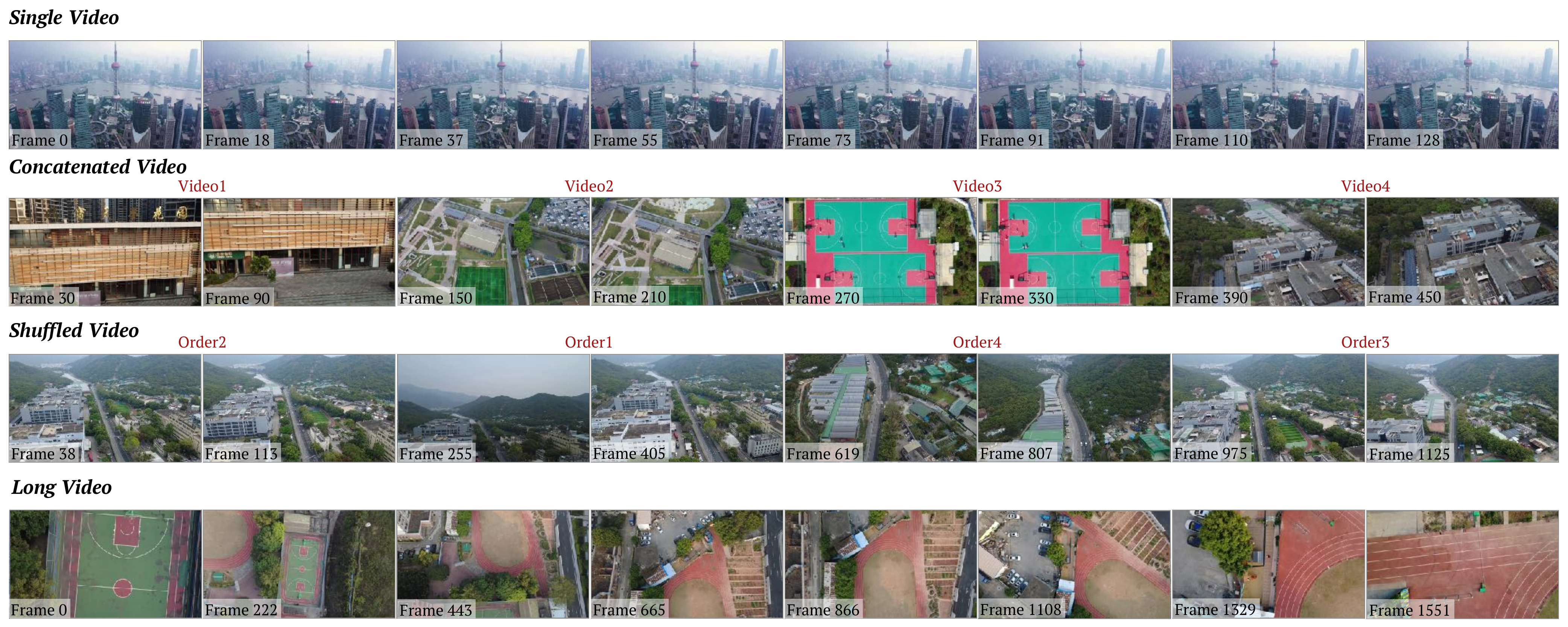}
\Description{A four-row figure illustrating the four processed video types in SIS-Bench. The rows show a Single Video, a Concatenated Video assembled from four short clips, a Shuffled Video formed by reordering four segments, and a Long Video kept in its original order. Frame indices and segment labels are overlaid.}
\caption{Representative examples of the four processed video types in SIS-Bench. From top to bottom: \emph{Single Video} keeps one original clip and is shown with uniformly sampled frames; \emph{Concatenated Video} stitches multiple short clips while preserving the order within each clip; \emph{Shuffled Video} reorders several segments from a longer trajectory, and the segment labels indicate the presented order; \emph{Long Video} preserves the original temporal order of a long flight. Frame indices are overlaid to show the sampling positions in the final processed video.}
\label{fig:data_construction_examples}
\end{figure}

Starting from the collected source data, we perform three processing operations, namely \emph{Converted}, \emph{Concatenated}, and \emph{Shuffled}, and retain part of the long videos in their original form. These steps finally yield four video types in the benchmark: \emph{Single Video}, \emph{Concatenated Video}, \emph{Long Video}, and \emph{Shuffled Video}. The definitions of these types are provided in the main paper (Sec.~3.1). Figure~\ref{fig:data_construction_examples} gives one representative example for each type.

\paragraph{VisDrone video conversion.}
For VisDrone, we convert raw frame sequences into videos at 15 FPS, i.e., every 15 frames correspond to 1 second of video time. This step turns frame-based data into temporally coherent video inputs that can be processed in the same way as the other benchmark sources.

\paragraph{AirScape clip concatenation.}
After the adjusted split, the AirScape benchmark pool contains 2067 short videos in total (923 from UrbanVideo, 444 from NAT2021, and 700 from WebUAV3M). We randomly sample 689 videos as \emph{Single Video} samples. The remaining 1378 videos are used to construct concatenated samples. Specifically, we randomly stitch 2--4 short clips into one video while keeping the proportions of 2-clip, 3-clip, and 4-clip compositions balanced, which finally yields 467 \emph{Concatenated Video} samples.

\paragraph{Long-video preservation.}
We keep a subset of long videos from UrbanVideo-Bench and VisDrone in their original temporal order, preserving continuous scene evolution and flight trajectories. This part finally provides 237 \emph{Long Video} samples.

\paragraph{Shuffled-video reordering.}
For shuffled-video construction, we use selected long UrbanVideo-Bench videos and adopt two different reordering strategies for two different tasks. For the \emph{Spatial Consistency} task, each video is split into four consecutive segments with unequal lengths. Specifically, we sample four random positive ratios, normalize them, and use them to partition the full video. The resulting four segments are then randomly permuted and concatenated into a new video. For the \emph{Spatio-temporal Consistency} task, each video is divided into four overlapping segments. The overlap is set to 3 seconds when the full video is shorter than 30 seconds and 5 seconds otherwise. After extracting the four overlapping clips, we randomly shuffle and concatenate them into a new video. Together, these two task-specific reordering strategies finally yield 253 \emph{Shuffled Video} samples.

\subsection{Summary of the processed benchmark videos}

\begin{figure}[H]
\centering
\includegraphics[width=\textwidth]{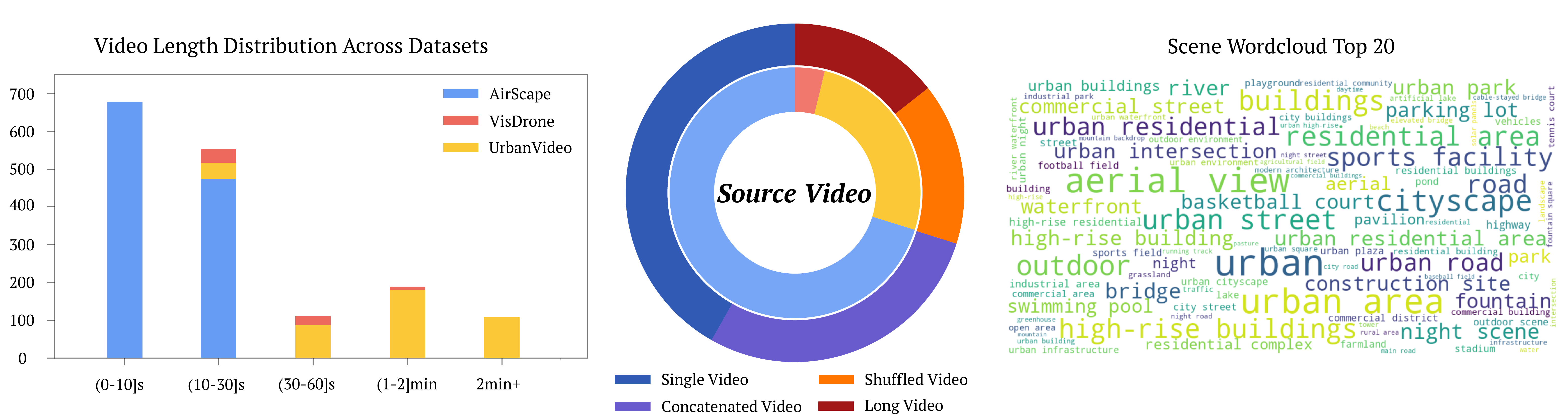}
\Description{A three-panel summary figure showing video length distributions across datasets, the composition of processed videos by type, and a word cloud of common scene descriptions.}
\caption{Statistics of the processed benchmark videos. Left: duration distribution across AirScape, UrbanVideo-Bench, and VisDrone, showing that AirScape mainly contributes short clips while UrbanVideo-Bench and VisDrone provide longer sequences. Center: composition of the processed video pool by construction type, including 689 \emph{Single Video} samples, 467 \emph{Concatenated Video} samples, 237 \emph{Long Video} samples, and 253 \emph{Shuffled Video} samples. Right: top-20 scene words, illustrating the diversity of benchmark environments.}
\label{fig:video_summary}
\end{figure}

After the above collection and processing steps, we obtain four video types in the benchmark: 689 \emph{Single Video} samples, 467 \emph{Concatenated Video} samples, 237 \emph{Long Video} samples, and 253 \emph{Shuffled Video} samples. Their duration distribution, type composition, and scene diversity are summarized in Figure~\ref{fig:video_summary}. Overall, the resulting videos span a wide temporal range from about 5 seconds to more than 2 minutes and cover diverse environments such as urban streets, residential areas, schools, parks, and waterfront regions.

\section{Task Design, Annotation, and QA Generation}

This section provides supplementary details on task formulation, metadata annotation, QA generation, expert review, and benchmark-level statistics.

\subsection{Task formulation details}
Following the benchmark taxonomy in the main paper, SIS-Bench evaluates 13 tasks across two dimensions and three cognitive levels. The task setup is tightly coupled with the four video types introduced in Section~\ref{sec:video_source_construction}, rather than being defined on a single uniform video format. Table~\ref{tab:task_breakdown} summarizes the resulting task organization, representative question patterns, video types, and the number of QA pairs for each task.

\paragraph{Spatial cognition: perception.}
The three perception tasks are all built on \emph{Single Video} inputs because they only require faithful understanding of one temporally coherent clip. \emph{Object Existence} asks whether a specified target appears in a short video and primarily evaluates object grounding and resistance to hallucination. \emph{Object Attribute} asks about a queried target's observable properties, such as color, shape, or textual content. \emph{Relative Direction} evaluates whether the model can identify the relative spatial relation between two landmarks or scene objects in the current view.

\paragraph{Spatial cognition: memory.}
The three memory tasks are built on \emph{Concatenated Video} inputs, where several short clips are stitched together while preserving the local order inside each clip. \emph{Landmark Order} asks the model to recover the order in which landmarks appear in a concatenated video. \emph{Landmark Recall} asks the model to retrieve the landmark observed immediately before a queried landmark, testing whether a coherent observation route has been formed. \emph{Positional Relationship} evaluates whether the model can retain and report the relative spatial relation between two landmarks observed across the video.

\paragraph{Spatial cognition: reasoning.}
\emph{Spatial Consistency} and \emph{Spatio-temporal Consistency} are both constructed from \emph{Shuffled Video} inputs, because these tasks are meant to probe reasoning under disrupted temporal order. \emph{Spatial Consistency} uses shuffled clips without reliable chronology and asks the model to reconstruct stable spatial relations from fragmented multi-view observations. \emph{Spatio-temporal Consistency} further requires the model to recover the original flight trajectory from shuffled clips with overlapping content, relying on shared landmarks and local motion continuity rather than the presented order.

\paragraph{Self-awareness: perception and memory.}
\emph{Action Recognition} uses \emph{Single Video} inputs because it focuses on the UAV's instantaneous motion state within one short clip. \emph{Action Sequence} and \emph{Action Recall} use \emph{Concatenated Video} inputs, since both tasks require the model to track action order and clip correspondence across multiple stitched segments. Specifically, \emph{Action Recognition} asks the model to identify the UAV's current flight action from a short clip. \emph{Action Sequence} requires the model to describe the ordered flight actions in a longer video. \emph{Action Recall} asks which action the UAV performed when it passed a particular scene element, location, or time point.

\paragraph{Self-awareness: reasoning.}
\emph{Action Prediction} and \emph{Path Planning} are built on \emph{Long Video} inputs because both tasks require access to continuous long-horizon motion context instead of fragmented clips. \emph{Action Prediction} asks the model to infer the likely consequence of a candidate action under the current state. \emph{Path Planning} provides a start view and a semantically specified target, and asks the model to generate an action sequence that reaches the target.

\begin{table*}[t]
\centering
\caption{Task breakdown of SIS-Bench. The table reports the two evaluation dimensions, three cognitive levels, question templates, video types, and QA counts for all 13 tasks.}
\label{tab:task_breakdown}
\small
\setlength{\tabcolsep}{4.5pt}
\renewcommand{\arraystretch}{1.24}
\begin{tabularx}{\textwidth}{>{\raggedright\arraybackslash}p{2.05cm}>{\centering\arraybackslash}p{1.45cm}>{\raggedright\arraybackslash}p{2.55cm}>{\centering\arraybackslash}p{1.9cm}>{\raggedright\arraybackslash}X>{\centering\arraybackslash}p{0.9cm}}
\hline
\textbf{Dimension} & \textbf{Level} & \textbf{Task} & \textbf{Video Type} & \textbf{Question Template} & \textbf{\#QA} \\
\hline
Spatial Cognition & Perception & Object Existence & Single & \textit{Which of the following objects does \textcolor{red}{\textbf{not appear}} in the video?} & 492 \\
Spatial Cognition & Perception & Object Attribute & Single & \textit{What is the \textcolor{red}{\textbf{[Attribute]}} of the \textcolor{red}{\textbf{[Object]}} in the video?} & 387 \\
Spatial Cognition & Perception & Relative Direction & Single & \textit{What object is located to the \textcolor{red}{\textbf{[Direction]}} of the \textcolor{red}{\textbf{[Reference Object]}}?} & 200 \\
Spatial Cognition & Memory & Landmark Order & Concatenated & \textit{What is the \textcolor{red}{\textbf{order of appearance}} of the landmarks in the video?} & 306 \\
Spatial Cognition & Memory & Landmark Recall & Concatenated & \textit{What landmark was observed \textcolor{red}{\textbf{immediately before}} the \textcolor{red}{\textbf{[Landmark]}}?} & 443 \\
Spatial Cognition & Memory & Positional Relationship & Concatenated & \textit{What object is located to the \textcolor{red}{\textbf{[Direction]}} of the \textcolor{red}{\textbf{[Reference Object]}}?} & 252 \\
Spatial Cognition & Reasoning & Spatial Consistency & Shuffled & \textit{The video is \textcolor{red}{\textbf{temporally shuffled}}. Assuming you are at [Obj A] facing [Obj B], what is to your \textcolor{red}{\textbf{[Direction]}}?} & 195 \\
Spatial Cognition & Reasoning & Spatio-temporal Consistency & Shuffled & \textit{The video is \textcolor{red}{\textbf{temporally shuffled}}. Starting from [Start Obj], what is the correct \textcolor{red}{\textbf{sequence}} of landmarks?} \newline \textit{In the reconstructed path, what landmark was passed \textcolor{red}{\textbf{between}} [Landmark A] and [Landmark B]?} & 241 \\
Self-Awareness & Perception & Action Recognition & Single & \textit{What \textcolor{red}{\textbf{flight action}} is the drone performing in the video?} & 686 \\
Self-Awareness & Memory & Action Sequence & Concatenated & \textit{What is the \textcolor{red}{\textbf{sequence}} of flight actions performed by the drone?} & 315 \\
Self-Awareness & Memory & Action Recall & Concatenated & \textit{In this \textcolor{red}{\textbf{stitched video}}, what action is being executed in the \textcolor{red}{\textbf{[N-th] clip}}?} & 804 \\
Self-Awareness & Reasoning & Action Prediction & Long & \textit{Starting from \textcolor{red}{\textbf{[Start Scene]}}, if the drone executes \textcolor{red}{\textbf{[Action Sequence]}}, what is its \textcolor{red}{\textbf{final position}}?} & 263 \\
Self-Awareness & Reasoning & Path Planning & Long & \textit{Starting from \textcolor{red}{\textbf{[Start Point]}}, what \textcolor{red}{\textbf{action sequence}} is needed to reach the \textcolor{red}{\textbf{[Target]}}?} & 272 \\
\hline
\end{tabularx}
\end{table*}

\subsection{Annotation details}
To support different task requirements across video types, we design three complementary annotation pipelines (A-C). 

\paragraph{Video-type assignment.}
As described in Sec.~B.1, we construct different task families based on distinct video types. 
Perception tasks are built from \emph{Single Video} samples, memory tasks are built from \emph{Concatenated Video} samples, and reasoning tasks are built from \emph{Shuffled Video} and \emph{Long Video} samples. This assignment follows the capability demands described in the main paper and avoids a uniform video format for all tasks.

\paragraph{Pipeline-A}
Pipeline-A covers the three tasks under self-awareness perception and memory, namely \emph{Action Recognition}, \emph{Action Sequence}, and \emph{Action Recall}. Only AirScape provides action annotations in the original data. Therefore, for these self-awareness tasks constructed from AirScape, we directly reuse the available action annotations and reorganize them into structured metadata. In this stage, we use GLM-4.6V-Flash-9B together with task templates to normalize the labels into fields such as action category, action order, and clip-level action correspondence.

\paragraph{Pipeline-B}
Pipeline-B covers the six tasks under spatial cognition perception and memory: \emph{Object Existence}, \emph{Object Attribute}, \emph{Relative Direction}, \emph{Landmark Order}, \emph{Landmark Recall}, and \emph{Positional Relationship}. For these tasks, we use model-assisted annotation to obtain structured scene metadata. The annotation process focuses on target existence, target attributes, OCR content, landmark identity, landmark order, and spatial relations. These automatically produced metadata are then manually filtered and corrected before later QA construction.

\begin{figure}[H]
\centering
\includegraphics[width=\textwidth]{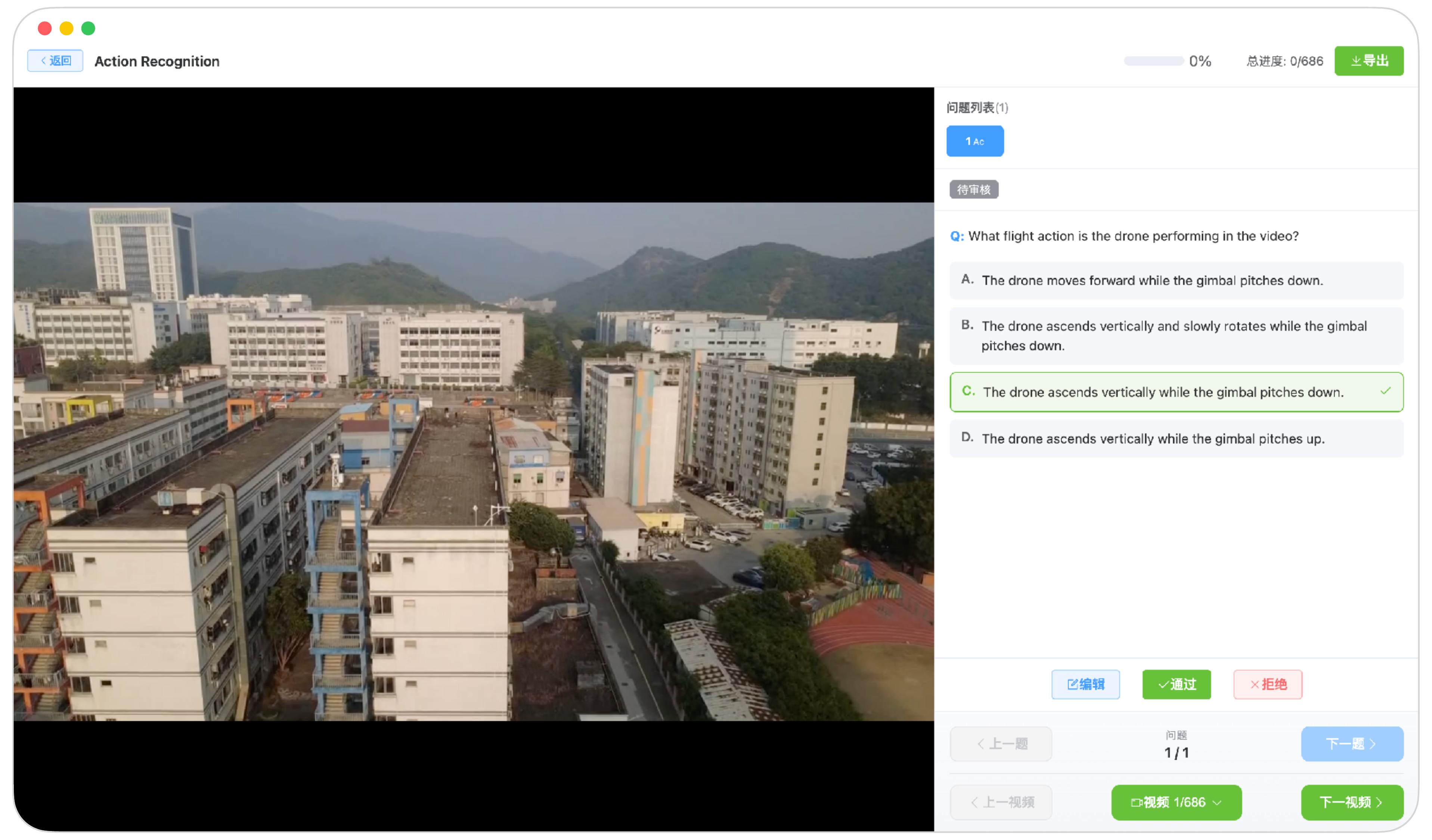}
\Description{The unified annotation and verification platform used in SIS-Bench, showing synchronized video inspection, question and option editing, and reviewer decisions.}
\caption{Unified annotation and verification platform used for expert annotation in Pipeline-C and subsequent dual-expert QA verification.}
\label{fig:annotation_verification_platform}
\end{figure}

\paragraph{Pipeline-C}
Pipeline-C covers the four reasoning tasks, including the two spatial-cognition reasoning tasks \emph{Spatial Consistency} and \emph{Spatio-temporal Consistency}, as well as the two self-awareness reasoning tasks \emph{Action Prediction} and \emph{Path Planning}. For reasoning tasks on \emph{Long Video} and \emph{Shuffled Video} inputs, automatic annotation is less reliable. We therefore adopt expert annotation to obtain complete metadata, including reference landmarks, route structure, candidate distractors, and grounded answers. These annotations are later converted into QA pairs with unified templates. The unified annotation and verification platform is shown in Figure~\ref{fig:annotation_verification_platform}.

\paragraph{Human annotation.}
Manual annotation and quality control were conducted by four annotators with master's-level training in EECS or AI, organized into two independent pairs. In Pipeline-B, they inspected the model-produced metadata and corrected object identities, attributes, landmark order, and spatial relations against the source videos. In Pipeline-C, they directly annotated the metadata required by the reasoning tasks because these samples involve longer temporal context and more complex route-level inference. The resulting QA pairs were subsequently checked through the dual-expert protocol detailed in Sec.~B.4.

\begin{figure}[H]
\centering
\includegraphics[width=\textwidth]{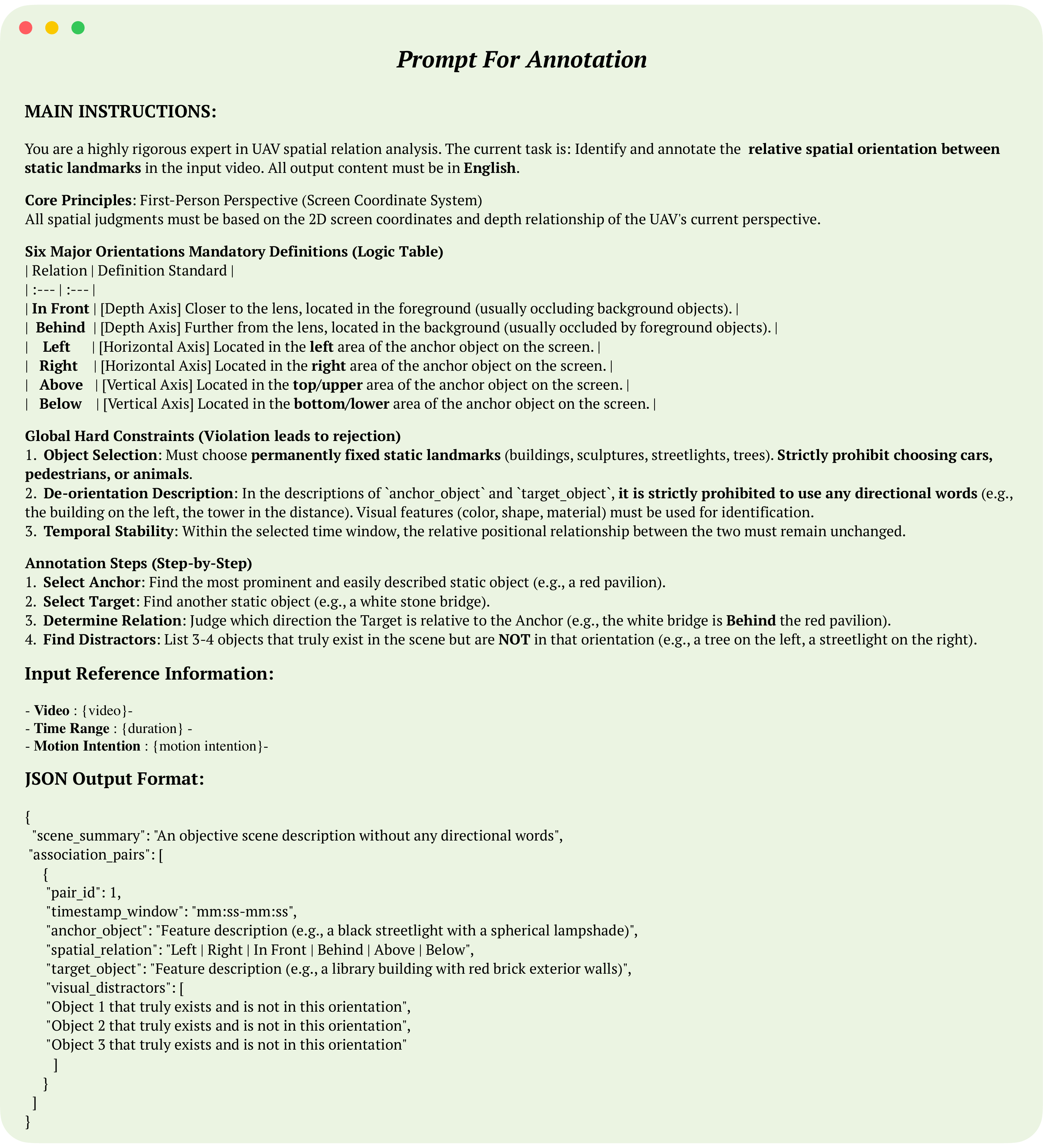}
\Description{An example annotation prompt template used in the benchmark annotation pipeline, illustrated with the Relative Direction task.}
\caption{Annotation template used during metadata construction, illustrated with the \emph{Relative Direction} task. We design task-specific templates for both the model-assisted and expert-assisted pipelines so that the collected metadata remains structured and directly usable for later QA construction.}
\label{fig:ann_template}
\end{figure}

\paragraph{Prompt design.}
Because annotation objectives vary across tasks, we design task-specific prompts following a unified principle: each prompt explicitly encodes the required cognitive target (e.g., perception, memory, or reasoning) and enforces structured, consistent outputs for downstream QA construction as the task requires. Accordingly, we use multiple templates rather than forcing all tasks into a single prompt.
In total, we design one template per task in Pipeline-B, 3 task templates in Pipeline-A, and 4 task templates in Pipeline-C. Figure~\ref{fig:ann_template} shows an example annotation template for the \emph{Relative Direction} task. 
In constructing prompts, we follow standard prompt design practices, including role definition, explicit task specification, constraint formulation, and structured output formats. Specifically, the prompt first explicitly defines a unified first-person screen-coordinate system so that terms such as \emph{front}, \emph{behind}, \emph{left}, \emph{right}, \emph{above}, and \emph{below} are interpreted consistently. Second, it restricts annotators to stable landmarks and requires object descriptions that do not leak directional cues. Third, it requests standardized JSON outputs so that the collected annotations can be directly reused for downstream QA construction and manual verification. Other task templates follow the same general philosophy while adapting the required fields and constraints to their own annotation targets.

\begin{figure}[H]
\centering
\includegraphics[width=\textwidth]{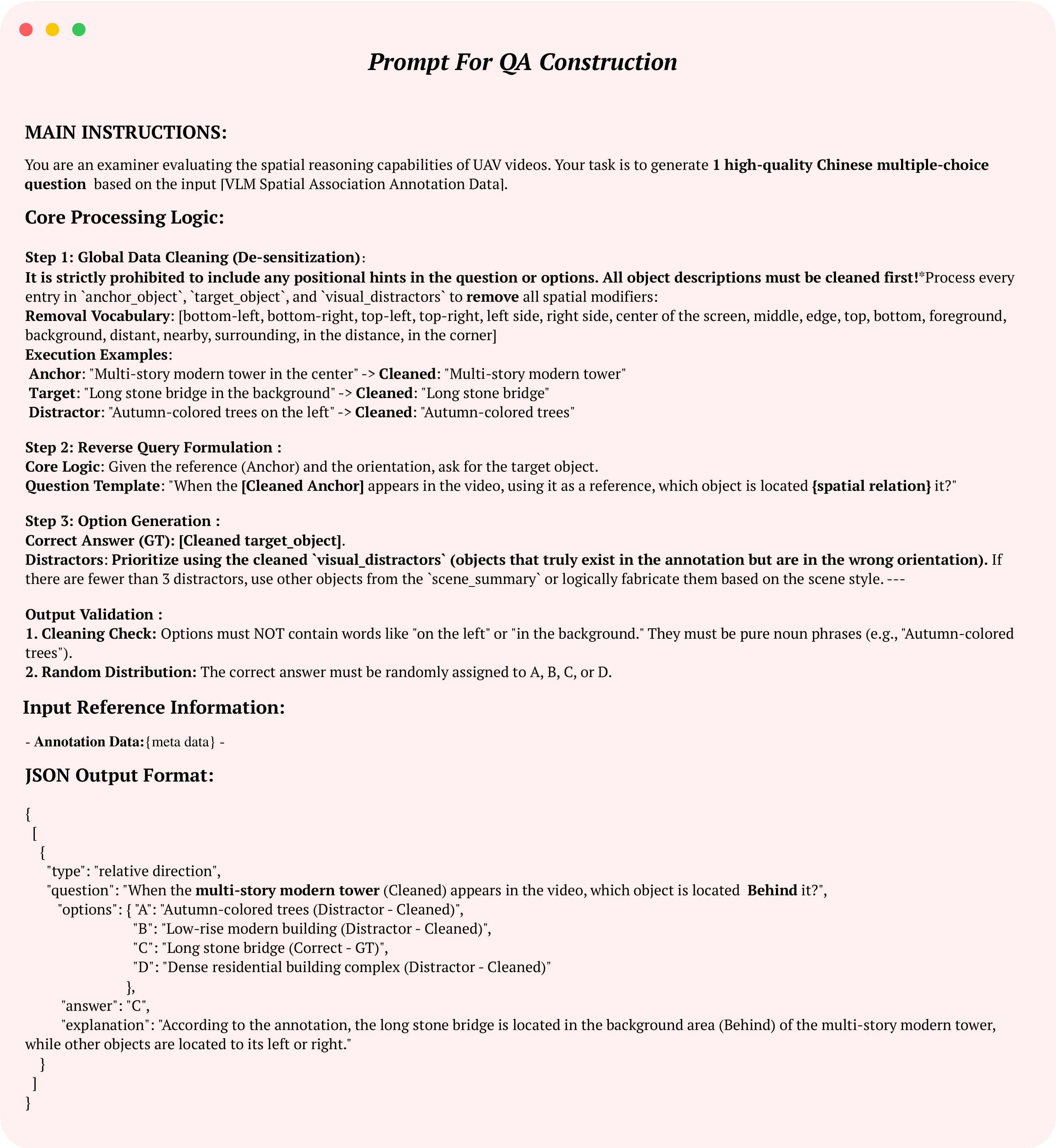}
\Description{An example QA-generation prompt template used to convert structured metadata into a multiple-choice question, illustrated with the Relative Direction task.}
\caption{QA-generation template, illustrated with the \emph{Relative Direction} task. We design one template for each of the 13 tasks and use multiple LLMs to improve the diversity of question phrasing and distractor construction while preserving answer uniqueness.}
\label{fig:qa_template}
\end{figure}

\subsection{QA generation details}

After collecting the required metadata, we generate multiple-choice QA pairs using task-specific templates.  Following the prompt design principles described in Sec.~B.2, we construct QA-generation prompts that emphasize explicit task specification, structured outputs, and consistent grounding in the underlying annotations. We use one QA-generation template for each of the 13 tasks. To increase diversity in wording and distractor construction, we use multiple LLMs, including Gemini 3 Flash, DeepSeek V3.2, MiMo-v2-Flash, and Qwen3-Max. The generation process preserves task-specific answer formats while ensuring that distractors remain visually plausible, semantically relevant, and non-ambiguous. The generated QA pairs are further checked to avoid lexical shortcuts, duplicate formulations, and answer leakage from the question text. 

Figure~\ref{fig:qa_template} shows an example QA-generation template for the \emph{Relative Direction} task.
We first clean the annotation text by removing explicit directional expressions and positional modifiers, so that the answer cannot be recovered by lexical overlap alone. We then transform the structured annotation into a question by taking the anchor object and the queried direction as conditions and asking for the target object. For option construction, the prompt prioritizes distractors that truly appear in the same scene but occupy incorrect orientations, and it randomizes the slot of the correct answer. Other task-specific QA templates follow the same principle of grounding questions in structured metadata while adapting the question form, answer type, and distractor strategy to the corresponding task.

\subsection{Dual-expert review details}

After large-scale QA generation, all candidate questions enter a dual-expert review stage. We use the unified annotation and verification platform shown in Figure~\ref{fig:annotation_verification_platform}. Two experts independently review each QA pair. A sample is accepted only when both reviewers approve it. Samples rejected by both reviewers are directly discarded. Samples with reviewer disagreement enter a discussion stage, after which they are either revised and retained or removed. During review, the experts focus on answer correctness, uniqueness of the correct option, grounding in the video evidence, distractor quality, and overall question validity.

In total, the review starts from 6,174 generated QA pairs. Among them, 1,944 are approved without modification, 2,912 are modified and retained after review, and 1,318 are rejected, resulting in the final 4,856 QA pairs. These outcomes correspond to approval, modification, and rejection rates of 31.5\%, 47.2\%, and 21.3\%, respectively. We use the modification rate as an operational indicator of review difficulty and reviewer dispute before resolution. Task-level statistics show that the more demanding reasoning tasks require substantially more intervention: 96.3\% of \emph{Action Prediction} samples are modified, while 41.2\% of \emph{Path Planning} samples are rejected.

\subsection{Benchmark summary}

SIS-Bench finally contains 13 tasks and 4856 multiple-choice QA pairs in total. The task distribution is shown in Table~\ref{tab:task_breakdown}. At the dimension level, spatial cognition contributes 2516 QA pairs and self-awareness contributes 2340. At the cognitive-level breakdown, perception contains 1765 QA pairs, memory 2120, and reasoning 971. Among the 13 tasks, \emph{Action Recall}, \emph{Action Recognition}, \emph{Object Existence}, and \emph{Landmark Recall} are the largest subsets, while the reasoning tasks remain comparatively compact because they require heavier expert annotation.

\subsection{Benchmark comparison}

\begin{table}[t]
\centering
\caption{Comparison of SIS-Bench with representative UAV/aerial benchmarks. We compare benchmark production mode, task focus, environment setting, and evaluation scale. SIS-Bench focuses on embodied UAV cognition in urban 3D aerial space using multi-source real-world UAV videos. All counts are rounded to one decimal place where applicable.}
\label{tab:benchmark_compare}
\small
\setlength{\tabcolsep}{4pt}
\renewcommand{\arraystretch}{1.18}
\begin{tabularx}{\textwidth}{
>{\raggedright\arraybackslash}p{2.2cm}
>{\raggedright\arraybackslash}p{1.9cm}
>{\raggedright\arraybackslash}X
>{\centering\arraybackslash}p{1.55cm}
>{\raggedright\arraybackslash}X
>{\raggedright\arraybackslash}p{1.85cm}
>{\centering\arraybackslash}p{1.45cm}
>{\centering\arraybackslash}p{1.25cm}}
\hline
\textbf{Benchmark} & \textbf{Production Mode} & \textbf{Task Focus} & \textbf{Embodied Agent} & \textbf{Environment Coverage} & \textbf{Motion Space} & \textbf{\#Source} & \textbf{\#Test Instances} \\
\hline
AerialVLN & Simulator-Based & UAV vision-language navigation & $\checkmark$ & Urban Outdoor Scenes & 3D Aerial Space & 8.4K paths & 25.3K \\
CityNav & Simulator-Based & Real-world aerial navigation & $\checkmark$ & Real Cities & 3D Aerial Space & 32.6K trajectories & 32.6K \\
OpenUAV & Simulator-Based & UAV visual-language navigation & $\checkmark$ & Complex Outdoor Scenes & 3D Aerial Space & 12.1K trajectories & 12.1K \\
UrbanVideo-Bench & Real-world & Aerial embodied cognition & $\checkmark$ & City & 3D Aerial Space & 1.5K videos & 5.2K \\
MM-UAVBench & Real-world & UAV MLLM evaluation & $\checkmark$ & Low-Altitude UAV Scenes & 3D Aerial Space & 1.5K videos + 2.8K images & 5.7K \\
\hline
Ours & Real-world & Self-in-space UAV cognition & $\checkmark$ & Diverse Urban UAV Scenes & 3D Aerial Space & 1.6K videos & 4.8K \\
\hline
\end{tabularx}
\end{table}

Table~\ref{tab:benchmark_compare} compares SIS-Bench with representative UAV/aerial benchmarks. The first three benchmarks in the table, namely AerialVLN, CityNav, and OpenUAV, are built with simulator-based production pipelines, which are well suited to large-scale generation of trajectories and test instances. In contrast, SIS-Bench is constructed from real-world UAV videos collected in common urban scenes. As a result, its design places greater emphasis on real-flight visual grounding and motion understanding than on benchmark scale alone.

Compared with the most relevant real-world UAV benchmarks, especially UrbanVideo-Bench and MM-UAVBench, SIS-Bench is comparable in overall scale while differing in benchmark organization. Specifically, it jointly evaluates \emph{spatial cognition} and \emph{self-awareness}, places all tasks within a unified perception--memory--reasoning hierarchy, and aligns task construction with four complementary video types, namely \emph{Single Video}, \emph{Concatenated Video}, \emph{Long Video}, and \emph{Shuffled Video}. This design allows SIS-Bench to provide a more fine-grained evaluation of embodied UAV cognition in real urban aerial environments.

\FloatBarrier
\section{Evaluation Protocol}

To reduce evaluation cost under limited computation, we preprocess all videos with adaptive frame sampling before inference. Our sampling strategy follows the internal practice used in Qwen-VL-style video processing and keeps at most 32 frames for each input video. Concretely, we use 2 FPS for videos shorter than 16 seconds, 1 FPS for videos between 16 and 32 seconds, and uniform 32-frame sampling for videos longer than 32 seconds. This strategy preserves denser temporal evidence for short videos while keeping the inference cost manageable for long videos.

We evaluate 20 open-source models and 6 proprietary models. The 20 open-source models are deployed on 4 RTX 4090 GPUs and inferred with the vLLM framework. For open-source evaluation, we use a unified setup with a 32,768-token context window, up to 128 generated tokens, at most 32 frames per video, no fixed image resizing, and pixel bounds from $256\times 28\times 28$ to $512\times 28\times 28$. The 6 proprietary models are evaluated through API-based inference under the same benchmark protocol.

\paragraph{Human baseline.}
We construct a task-balanced human-study subset by sampling 80 QA pairs from each of the 13 tasks, for 1,040 questions in total. Six participants with master's-level training in EECS or AI, none of whom participated in benchmark annotation or verification, independently answer all 1,040 questions, yielding 6,240 individual responses. Participants are allowed to replay each video while answering. The reported human accuracy of 91.7\% is the average of the six participants' individual accuracies.

\section{Deeper Diagnosis}
\label{sec:deeper_diagnosis}

Beyond aggregate accuracy, we use controlled evaluations, matched model comparisons, an error taxonomy, and instance-level failures to diagnose the observed trends and remaining reasoning limitations.

\subsection{Robustness controls and model-type analysis}

\begin{figure}[H]
\centering
\includegraphics[width=\textwidth]{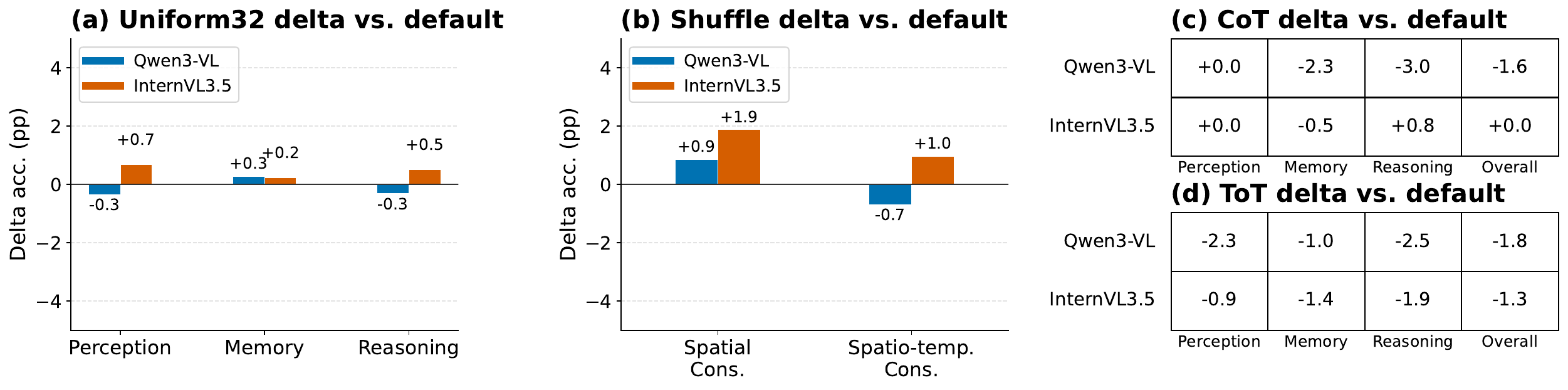}
\Description{Robustness controls showing accuracy changes under uniform 32-frame sampling, repeated shuffled ordering, Chain-of-Thought prompting, and Tree-of-Thought prompting.}
\caption{Controls for potential confounding factors. (a) Uniform 32-frame sampling. (b) Repeated random shuffling. (c) Chain-of-Thought prompting. (d) Tree-of-Thought prompting. Values report accuracy changes relative to the default evaluation.}
\label{fig:robustness_controls}
\end{figure}

\paragraph{Confounding-factor controls.}
We conduct four controls to test whether the perception$>$memory$>$reasoning trend is caused by input artifacts rather than model capability. First, beyond the default 2~FPS sampling with at most 32 frames, we evaluate uniform 32-frame sampling to equalize the frame budget across video lengths. The level-wise changes remain within 0.7 percentage points for both Qwen3-VL-8B and InternVL3.5-8B, leaving the trend unchanged. Second, for shuffled-video tasks, we repeat random shuffle-and-concatenate three times and average the results; the changes remain marginal, indicating that the conclusion is not driven by one favorable or unfavorable ordering. Third, we compare direct answering with Chain-of-Thought (CoT) and Tree-of-Thought (ToT) prompting. As shown in Figure~\ref{fig:robustness_controls}, neither strategy produces consistent gains: CoT changes reasoning accuracy by $-3.0$ points for Qwen3-VL and $+0.8$ points for InternVL3.5, while ToT reduces reasoning accuracy for both models. Finally, human participants answer the same concatenated, long, and shuffled inputs and remain near 90\% accuracy at every cognitive level. Thus, neither video construction nor intrinsic task solvability explains the much steeper model degradation; the trend primarily reflects model-side limitations.

\paragraph{Reasoning-oriented models.}
We further compare matched model pairs in Table~\ref{tab:main_results}. Switching from Qwen3-VL-8B-Instruct to Qwen3-VL-8B-Thinking improves weighted reasoning accuracy by 2.5 points, while switching from MiMo-VL-7B-SFT to MiMo-VL-7B-RL yields a smaller 0.6-point gain. The improvements are most visible on temporal action reasoning, showing that explicit reasoning training can help selected tasks. However, the gains are task-dependent, Path Planning can still decrease, and the overall perception$>$memory$>$reasoning ordering remains. Together with the prompting controls, these matched comparisons indicate that stronger language-side reasoning helps but does not remove the underlying bottlenecks in temporal grounding, spatial consistency, and UAV ego-motion estimation.

\subsection{Error taxonomy and cognitive bottlenecks}

\begin{figure}[H]
\centering
\includegraphics[width=0.78\textwidth,trim=0 0 1470 0,clip]{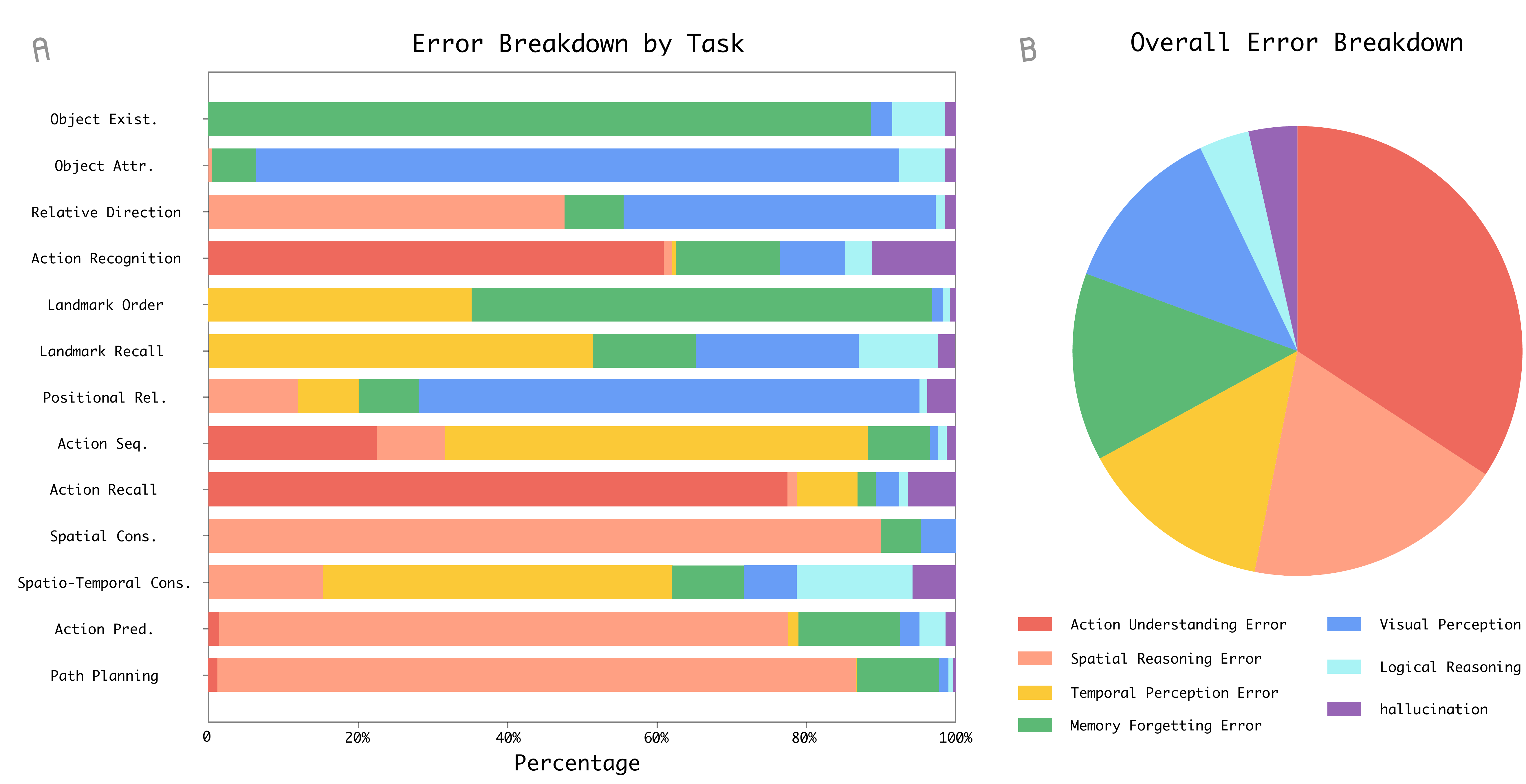}
\Description{Error analysis over incorrect model responses, including task-wise stacked error distributions and an overall error breakdown.}
\caption{Diagnostic error breakdown over 7,987 incorrect model responses. Left: task-wise distributions of dominant error types. Right: overall error distribution.}
\label{fig:error_breakdown}
\end{figure}

Following the diagnostic spirit of VSI-Bench~\cite{VSI-Bench}, we analyze 7,987 incorrect responses with chain-of-thought traces from four representative models: Qwen3-VL-8B-Thinking, Step3-VL-10B, Ovis2.5-9B, and GLM-4.1V-9B-Thinking. We use eight predefined categories covering visual and temporal perception, spatial and logical reasoning, action understanding, memory forgetting, instruction misunderstanding, and hallucination. Figure~\ref{fig:error_breakdown} reports the task-wise distribution and the overall breakdown of the dominant error types.

The two largest categories are \emph{Action Understanding Error} (33.0\%) and \emph{Spatial Reasoning Error} (18.2\%). Thus, the failures cannot be attributed only to missed objects or poor local visual recognition. Many models identify the relevant entities but still fail to infer the UAV's ego-motion, distinguish rotation from translation, or maintain a consistent egocentric orientation. The task-wise distribution reinforces this interpretation: action-related tasks are dominated by action-state errors, whereas relative-direction, spatial-consistency, and path-planning tasks exhibit strong spatial-reasoning failures.

These patterns provide a causal interpretation of the aggregate trends. Spatial perception tasks can often be solved from local object appearance and static relations, while self-awareness tasks require the model to estimate and track the UAV's own motion state. Moving from perception to memory further adds cross-time binding, and reasoning requires joint temporal reconstruction, spatial consistency, and action-consequence inference. SIS-Motion supports this diagnosis: relative to the zero-shot backbone, it improves all nine perception and memory tasks, but its gains over visual-only SFT remain limited and inconsistent on the four reasoning tasks in Table~\ref{tab:ablation}. Explicit local motion cues therefore reduce early-stage perception and tracking errors but do not resolve long-horizon temporal and ego-motion reasoning.

\FloatBarrier
\subsection{Below-random Path Planning performance}

We first clarify that the 20.6\% SFT result, the 25.2\% random baseline, and the 23.5\% SIS-Motion result correspond to \emph{Path Planning}, not \emph{Action Prediction}. For \emph{Action Prediction}, the task-specific random baseline is 24.9\%, while visual-only SFT reaches 45.2\% and SIS-Motion reaches 44.1\% in Table~\ref{tab:ablation}. The below-random observation is therefore specific to Path Planning rather than a general failure across both self-awareness reasoning tasks.

This limitation is consistent with the training scope of SIS-Motion-54K. The training set supervises the nine perception and memory tasks but contains no planning-oriented examples. Accordingly, SFT improves all nine supervised capability categories, while Path Planning remains near chance. Path Planning requires goal-conditioned, long-horizon inference over scene constraints, action consequences, and the UAV's evolving pose; local optical-flow cues alone do not provide this capability. Moreover, errors are systematic rather than uniformly random: the Path Planning example in Table~\ref{tab:error_cases} mirrors the required left--right turn and repeatedly selects a plausible but directionally incorrect route. We therefore treat the below-random result as evidence of an unresolved planning bottleneck and avoid claiming that motion-aware training improves high-level reasoning in general.

\subsection{Representative failure cases}

\begin{figure}[!t]
\centering
\includegraphics[width=\textwidth]{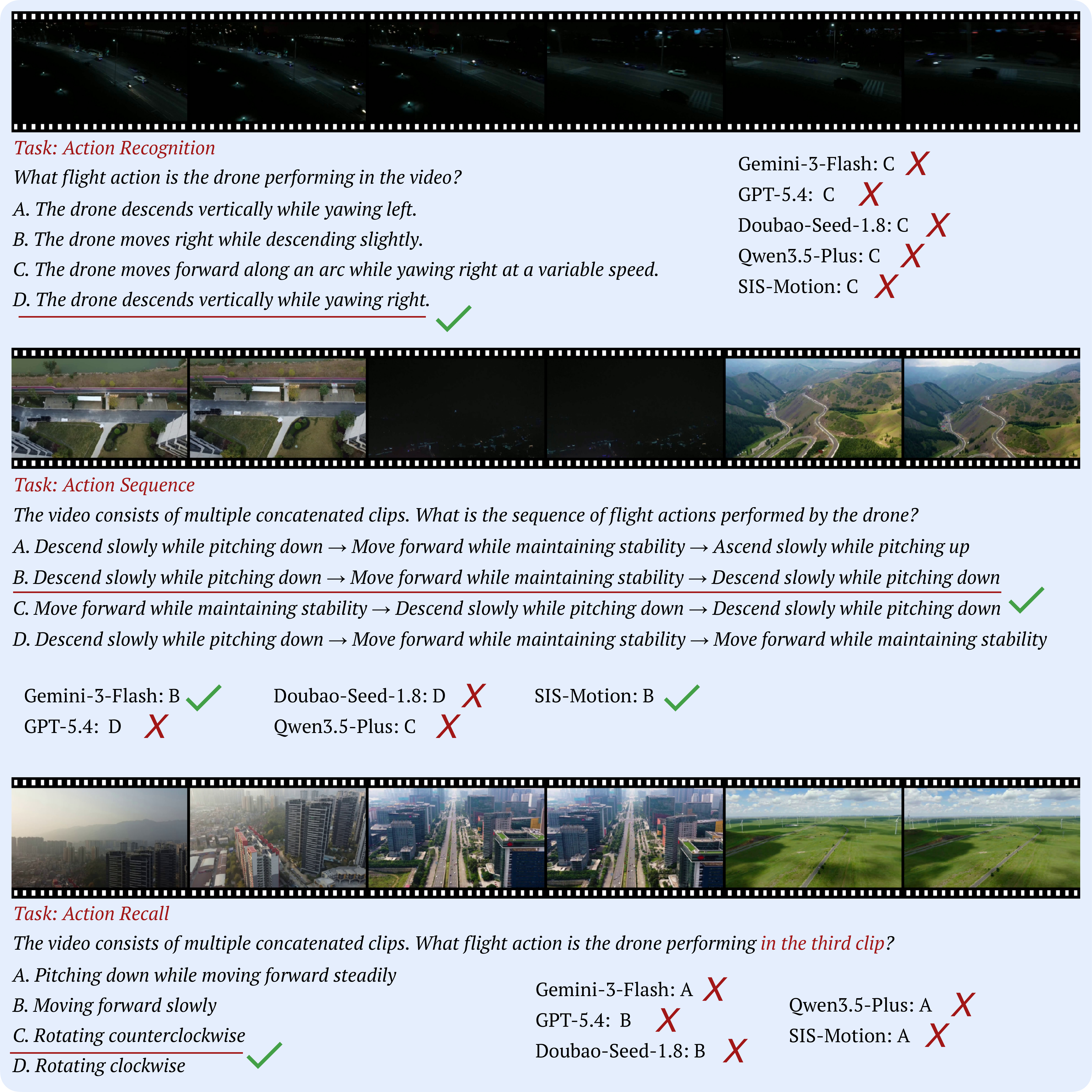}
\caption{Representative failures in \emph{Action Recognition} (top), \emph{Action Sequence} (middle), and \emph{Action Recall} (bottom), highlighting weak motion grounding, temporal binding, and rotation--translation discrimination.}
\label{fig:error_case_example}
\end{figure}

Table~\ref{tab:error_cases} lists one representative error case for each of the 13 tasks, so that readers can inspect a concrete failure mode for every task category. To provide a more detailed discussion, Figure~\ref{fig:error_case_example} further zooms in on three action-related tasks, namely \emph{Action Recognition}, \emph{Action Sequence}, and \emph{Action Recall}. These three examples are particularly informative because they require the model to infer the UAV's own motion state, maintain clip-level temporal structure, and distinguish fine-grained ego-motion patterns rather than relying only on static scene semantics.

\begingroup
\small
\setlength{\tabcolsep}{5pt}
\renewcommand{\arraystretch}{1.22}
\newcolumntype{L}[1]{>{\raggedright\arraybackslash}p{#1}}
\begin{longtable}{L{2.5cm}L{4.35cm}L{2.9cm}L{4.35cm}}
\caption{Representative error cases on SIS-Bench from Qwen3-VL-8B-Thinking across all 13 tasks. We provide one example for each task to illustrate typical failure modes in perception, memory, reasoning, and self-awareness. The last column gives an abbreviated summary of the model's original response.}
\label{tab:error_cases} \\
\hline
\textbf{Question Category} & \textbf{Question} & \textbf{Ground Truth} & \textbf{Video-LLM Answer} \\
\hline
\endfirsthead
\hline
\textbf{Question Category} & \textbf{Question} & \textbf{Ground Truth} & \textbf{Video-LLM Answer} \\
\hline
\endhead
Object Existence & Which object does not appear in the video? Options: A harvester; B green trailer; C stacked hay bales; D blue water truck. & D: blue water truck & C: stacked hay bales. The model confuses crop appearance with hay bales and overlooks that the truck never appears. \\
Object Attribute & What is the color of the village residential buildings? Options: A light pink and white; B mixed pink and white; C blue and white alternating; D gray and black. & B: mixed pink and white & A: light pink and white. It over-focuses on one tone and misses that the buildings appear as a mixed pink-white group. \\
Relative Direction & What object is located to the rear of the dark column-shaped parking sign? Options: A blue sedan; B red sedan; C white sedan; D white street lamp. & C: white sedan & A: blue sedan. The model anchors on the most salient nearby car and fails to judge the correct depth relation behind the sign. \\
Landmark Order & What is the chronological order of these landmarks: canal, red-white communication tower, residential complex, and blue glass skyscraper? & B: canal -> tower -> residential complex -> skyscraper & C: canal -> residential complex -> tower -> skyscraper. It swaps the tower and the residential complex. \\
Landmark Recall & What landmark was observed immediately before the glass curtain wall office building? Options: arched bridge, beige commercial building, gray-white office building, red circular stadium. & C: gray-white office building & B: beige commercial building. The model uses coarse clip grouping and misses the directly preceding landmark transition. \\
Positional Relationship & In clip 4, what lies to the right of the white glass-curtain-wall high-rise? Options include a red-spire white building, a red multi-story building, a transmission tower, and a blue building. & B: red multi-story building & A: white multi-story building with red spire. It confuses the referred clip and picks a nearby but incorrect landmark. \\
Spatial Consistency & If the UAV hovers above the red-top pavilion and faces the circular blue fountain, what landmark is on the right? Options: irregular pond, signal tower, blue rectangular pool, red multi-story building. & A: irregular pond & C: blue rectangular pool. The model reconstructs the scene with an incorrect egocentric orientation and places the pool on the wrong side. \\
Spatio-temporal Consistency & The flight starts from the rectangular swimming pool. What is the correct landmark sequence? Options include pool -> basketball court -> fountain and reordered alternatives. & A: pool -> basketball court -> fountain & C: basketball court -> pool -> fountain. It trusts the apparent opening clip more than the instruction and reconstructs the route in the wrong order. \\
Action Recognition & What flight action is the drone performing? Options: rotate backward in place, move backward steadily, move backward slowly, move forward steadily. & B: move backward steadily & D: move forward steadily. The model interprets the increasing scale of buildings as forward motion and misses the actual backward movement cue. \\
Action Sequence & What is the action sequence in the concatenated video? The key difference is whether the middle segment yaws left or right. & A: forward + pitch down -> forward + yaw left -> forward + pitch down & D: forward + pitch down -> forward + yaw right -> forward + pitch down. It captures the three-stage structure but reverses the middle turning direction. \\
Action Recall & What flight action is the drone performing in the second clip? Options include rotating left while moving forward and moving forward while pitching down. & C: moving forward while pitching down & A: rotating left while moving forward. The model over-interprets the viewpoint shift as yaw instead of recognizing the forward motion with pitch change. \\
Action Prediction & Starting above the red tennis court, if the drone pitches up, yaws right 90 degrees, flies forward, and descends, where does it reach? Options include greenhouse, office building, basketball court, and runway. & C: basketball court with dark blue floor & B: high-rise office building with white glass curtain wall. It follows the turn toward distant buildings and misses the intended landing target. \\
Path Planning & Starting above the red tennis court, what action sequence reaches the blue-floor basketball court? The options mainly differ in whether the UAV should yaw left or right. & B: pitch up -> yaw right 45 -> fly forward -> descend & A: pitch up -> yaw left 45 -> fly forward -> descend. It mirrors the left-right orientation and plans the route in the wrong direction. \\
\hline
\end{longtable}
\endgroup

\noindent \textbf{Motion ambiguity under weak nighttime evidence.}
The first example is a nighttime video with sparse illumination and weak texture. The correct answer is option D, \emph{the drone descends vertically while yawing right}, but all compared general-purpose models in Figure~\ref{fig:error_case_example} choose option C, namely \emph{the drone moves forward along an arc while yawing right at a variable speed}. The model must infer ego-motion from subtle frame changes: only a few road lights and vehicle traces are visible, while most background structure is dark and low-contrast. As a result, the models appear to over-rely on a small number of salient light trajectories and interpret the observed change as forward curved motion. What they miss is the global motion pattern of the camera: the overall viewpoint change is more consistent with vertical descent combined with rightward yaw than with sustained forward translation. This case suggests that under weak illumination, current MLLMs still lack robust motion grounding and can be easily misled by sparse but visually salient appearance changes.

\noindent \textbf{Failure to preserve segment boundaries and order.}
The second example is a concatenated video composed of three clips. The correct answer is option B, namely \emph{descend slowly while pitching down} $\rightarrow$ \emph{move forward while maintaining stability} $\rightarrow$ \emph{descend slowly while pitching down}. This requires detecting three segments, assigning an action to each, and preserving their order. Figure~\ref{fig:error_case_example} reveals two typical errors. Models answering option D correctly identify the first descent and the middle forward motion, but they fail to detect that the last clip returns to a descending state and instead treat it as a continuation of forward flight. Models answering option C make an earlier mistake by shifting the forward-motion interpretation to the first clip, which then corrupts the whole sequence. In other words, the main failure is not total ignorance of local motion cues, but weak boundary detection and weak temporal binding across stitched clips. Once one segment is misread, the full action chain becomes inconsistent.

\noindent \textbf{Confusion between rotation and translation in the queried clip.}
The third example also comes from a concatenated video, but now the question asks for the action in the \emph{third clip}. The correct answer is option D, \emph{rotating clockwise}. To solve this case, the model must complete two steps correctly: it must first localize the queried third segment within the multi-clip video, and then distinguish the motion type inside that segment. In Figure~\ref{fig:error_case_example}, the compared models instead predict either option A, \emph{pitching down while moving forward steadily}, or option B, \emph{moving forward slowly}. This pattern is revealing. The models do capture that the viewpoint is changing, but they explain that change as translation rather than in-place rotation. Visually, the queried clip contains substantial heading change while the scene layout remains comparatively stable, which is more consistent with rotation than with forward advance. However, models that rely mainly on frame appearance differences tend to map any obvious view change to ``moving forward'' or ``pitching down,'' without checking whether the background geometry supports genuine translational displacement. This error shows that fine-grained ego-motion discrimination remains difficult even when the relevant clip has already been specified in the question.

Taken together, these three examples show that current MLLMs do not simply fail because the scene is hard to see; they more fundamentally struggle to convert visual evidence into a stable estimate of the UAV's own motion state. The errors therefore arise at multiple levels, including motion grounding under weak visibility, segment-level temporal organization, and fine-grained discrimination between different ego-motion types.

\FloatBarrier
\section{SIS-Motion Training Details and Ablations}

\subsection{SIS-Motion-54K construction}

To evaluate the effectiveness of motion-aware (self-related) modeling, we further construct a training corpus named \emph{SIS-Motion-54K} in Sec.~5 . Its overall construction pipeline follows the benchmark pipeline described above, including split control, metadata annotation, and instruction generation, but all training data are drawn from the \emph{training} side of the adjusted AirScape split introduced in Section~1. In this way, the training corpus remains strictly separated from benchmark construction.

More specifically, SIS-Motion-54K is built from the adjusted AirScape training pool (0 UrbanVideo clips, 1349 NAT2021 clips, and 7596 WebUAV3M clips) and uses the same task templates as the benchmark pipeline. The corpus contains three complementary data forms.

\paragraph{MCQ data.}
The first part is a multiple-choice instruction set covering the nine perception and memory tasks. This part exposes the model to grounded recognition, clip-level memory, and cross-segment reasoning cues under the same task definitions used in benchmark evaluation, thereby improving task alignment between training and testing.

\paragraph{OpenQA data.}
The second part is obtained by rewriting the MCQ samples into open-ended question--answer pairs. This conversion removes explicit answer options and encourages the model to produce grounded responses directly from video evidence, which helps reduce reliance on option matching.

\paragraph{Description data.}
The third part contains descriptive supervision. For action-intent descriptions, we directly use the original AirScape action annotations. For scene-content descriptions, we reuse the same template-based annotation strategy as in benchmark construction to produce structured long-form descriptions of landmarks, scene layout, and motion cues. This part is designed to preserve and strengthen the model's long-text generation ability rather than restricting training to short-form multiple-choice outputs.

\subsection{Training details}

Because our computation budget is limited, all baseline, SIS-Motion, and ablation experiments are conducted with parameter-efficient LoRA fine-tuning. Training and evaluation are both carried out on 4 RTX 4090 GPUs, and the evaluation protocol follows Sec.~3 exactly.

\paragraph{Common settings.}
Unless otherwise specified, we use one training epoch, BF16 training, maximum sequence length 2048, gradient checkpointing, and 4 dataloader workers. The per-device batch size is 1 with gradient accumulation of 8. We set the base learning rate to $1\times10^{-5}$, use a cosine scheduler with warmup ratio 0.03 and weight decay 0.01, and adopt LoRA with rank 32, alpha 64, and dropout 0.05. For video inputs, we sample at 2 FPS, keep 8--32 frames per sample, and constrain the spatial resolution to the range from $128\times 28\times 28$ to $512\times 28\times 28$ pixels.

\paragraph{Baseline fine-tuning.}
For the baseline model, we freeze the vision encoder and fine-tune the remaining trainable components on SIS-Motion-54K for one epoch. Checkpoints are saved every 500 steps, with at most 3 checkpoints retained.

\paragraph{SIS-Motion fine-tuning.}
For SIS-Motion, we freeze both the vision encoder and the motion encoder, keep LoRA adaptation enabled, and tune the multimodal connector for one epoch. In this setting, the projector learning rate is set to $2\times10^{-5}$. We save checkpoints every 100 steps and retain at most 3 checkpoints.

\subsection{Ablation settings and results}

We conduct two groups of ablation experiments to analyze the contribution of the motion modeling design. First, we replace the optical-flow estimator with RAFT, Sea-RAFT, and MemFlow to examine whether the gains are tied to a specific motion extraction choice. Second, we replace the 3B Qwen2.5-VL backbone with a 7B backbone to study the effect of backbone capacity. Table~\ref{tab:exp_results} summarizes the full comparison among vanilla backbones, SFT baselines, motion variants, and backbone scaling.

Overall, the ablation results lead to two consistent conclusions. First, all motion-enhanced variants outperform the corresponding vanilla 3B baseline, indicating that the improvement does not depend on a single optical-flow estimator and that introducing explicit motion cues is itself beneficial. Second, the 7B variant further improves over the 3B setting (from 69.1 to 76.9) and remains stronger than the corresponding vanilla backbone (from 73.1 to 76.9), showing that the proposed design is compatible with backbone scaling. These observations are consistent with the conclusions reported in the main paper and provide additional evidence that the motion modeling strategy contributes meaningfully beyond a single implementation choice.

\setlength{\tabcolsep}{3pt}
\begin{table*}[t]
\centering
\caption{Experimental results on SIS-Bench. Accuracy (\%) across 13 tasks for vanilla Qwen2.5-VL backbones, SFT baselines, motion-modeling variants, and backbone scaling. MOFNet is the motion module used in SIS-Motion.}
\label{tab:exp_results}
\scriptsize
\renewcommand{\arraystretch}{1.05}
\resizebox{\textwidth}{!}{
\begin{tabular}{l|ccc|ccc|ccc|cc|c|cc|cc}
\hline
& & & & \multicolumn{8}{c|}{\textcolor[HTML]{675BC6}{\textbf{Spatial Cognition}}} & \multicolumn{5}{c}{\textcolor[HTML]{FF7F00}{\textbf{Self-Awareness}}} \\
& & & & \multicolumn{3}{c|}{\cellcolor[HTML]{EBF4E2}Perception} & \multicolumn{3}{c|}{\cellcolor[HTML]{FDF0EF}Memory} & \multicolumn{2}{c|}{\cellcolor[HTML]{E5EEFD}Reasoning} & \cellcolor[HTML]{EBF4E2}Perc. & \multicolumn{2}{c|}{\cellcolor[HTML]{FDF0EF}Memory} & \multicolumn{2}{c}{\cellcolor[HTML]{E5EEFD}Reasoning} \\
\textbf{Model} & \rotatebox{90}{\textbf{Spatial Avg}} & \rotatebox{90}{\textbf{Self Avg}} & \rotatebox{90}{\textbf{Overall}} & \rotatebox{90}{\textit{Obj Exist}} & \rotatebox{90}{\textit{Obj Attr}} & \rotatebox{90}{\textit{Rel Dir}} & \rotatebox{90}{\textit{Land Order}} & \rotatebox{90}{\textit{Land Recall}} & \rotatebox{90}{\textit{Pos Rel}} & \rotatebox{90}{\textit{Spat Consist}} & \rotatebox{90}{\textit{ST Consist}} & \rotatebox{90}{\textit{Act Recog}} & \rotatebox{90}{\textit{Act Seq}} & \rotatebox{90}{\textit{Act Recall}} & \rotatebox{90}{\textit{Act Pred}} & \rotatebox{90}{\textit{Path Plan}} \\
\hline
\multicolumn{17}{l}{\cellcolor[HTML]{ECF4FF}\textit{Vanilla Backbones}} \\
Qwen2.5-VL-3B & 65.8 & 40.5 & 53.6 & 92.7 & 48.8 & 64.5 & 90.8 & 51.9 & 62.7 & 57.4 & 42.7 & 37.9 & 62.5 & 35.6 & 53.6 & 23.2 \\
Qwen2.5-VL-7B & 69.9 & 40.7 & 55.8 & 96.5 & 71.1 & 69.0 & 73.2 & 54.2 & 69.0 & 67.7 & 41.5 & 43.3 & 55.6 & 29.2 & 60.8 & 31.6 \\
\hline
\multicolumn{17}{l}{\cellcolor[HTML]{ECF4FF}\textit{SFT Baselines}} \\
Qwen2.5-VL-3B SFT & 72.0 & 60.3 & 66.4 & 96.1 & 67.7 & 66.0 & 93.8 & 63.4 & 69.8 & 57.4 & 36.9 & 88.0 & 80.6 & 47.1 & 45.2 & 20.6 \\
Qwen2.5-VL-7B SFT & 79.3 & 66.5 & 73.1 & 96.7 & 82.9 & 73.5 & 93.8 & 76.1 & 77.0 & 62.1 & 46.9 & 90.8 & 85.7 & 54.4 & 58.9 & 26.1 \\
\hline
\multicolumn{17}{l}{\cellcolor[HTML]{ECF4FF}\textit{Motion Variants on Qwen2.5-VL-3B}} \\
RAFT + 3B & 73.8 & 59.9 & 67.1 & 96.5 & 72.4 & 67.0 & 92.2 & 68.6 & 66.7 & 56.9 & 42.7 & 87.9 & 78.7 & 47.8 & 41.4 & 21.0 \\
Sea-RAFT + 3B & 74.6 & 61.0 & 68.1 & 95.3 & 70.3 & 67.5 & 94.8 & 70.2 & 71.4 & 58.5 & 44.4 & 88.2 & 78.7 & 49.0 & 45.6 & 22.4 \\
MemFlow + 3B & 73.9 & 62.2 & 68.2 & 96.1 & 70.3 & 65.5 & 93.5 & 66.4 & 71.8 & 59.5 & 43.6 & 88.6 & 81.3 & 51.6 & 44.5 & 22.1 \\
MOFNet + 3B & 74.2 & 63.7 & 69.1 & 96.5 & 71.1 & 66.0 & 93.8 & 68.2 & 73.0 & 60.0 & 39.8 & 88.1 & 81.3 & 55.9 & 44.1 & 23.5 \\
\hline
\multicolumn{17}{l}{\cellcolor[HTML]{ECF4FF}\textit{Backbone Scaling}} \\
MOFNet + 7B & 79.9 & 73.6 & 76.9 & 97.6 & 82.2 & 75.0 & 95.8 & 75.8 & 77.8 & 64.6 & 46.5 & 91.3 & 87.6 & 73.4 & 56.7 & 29.8 \\
\hline
\end{tabular}
}
\end{table*}

\section{Downstream Navigation Transfer}

To further evaluate the practical utility of the model in UAV decision-making, we construct a downstream path-planning task based on OpenUAV. Since OpenUAV is originally designed for vision-language navigation and provides target pose annotations rather than discrete language outputs, we adapt it into a form suitable for multimodal large-model evaluation. Two typical navigation scenarios are depicted as Figure \ref{fig:downstream_task}.

\paragraph{Data collection and processing.}
We extract UAV flight trajectories from OpenUAV and collect the front-view image at each time step, which is then organized into a video sequence. Based on the relative coordinate sequence, we derive the flight operation at each step. Concretely, an altitude increase larger than 5 meters is labeled as \emph{take off}, while an altitude decrease larger than 5 meters is labeled as \emph{landing}. We then compute the angle between adjacent horizontal displacement vectors: if the turning angle falls between $25^\circ$ and $120^\circ$, the step is labeled as \emph{left turn} or \emph{right turn}; otherwise, it is treated as \emph{cruise}. Consecutive steps with the same operation are merged into one flight segment, after which we compute the segment-level flight distance, altitude change, and turning angle to obtain a rule-based path description.

\paragraph{Multiple-choice question construction.}
To fit the evaluation format used in this paper, we convert the collected videos and rule-based path descriptions into four-choice questions. We use an LLM to rewrite the rule descriptions into natural question wording, then use the true path as the correct option and construct three distractors. The distractors are generated through direction reversal (e.g., replacing left with right), large distance perturbation, insertion or deletion of operation stages, and incorrect altitude changes, so that the options remain challenging but unambiguous.

\paragraph{Test-set scale.}
We sample 22 simulated scene maps and finally construct a downstream test set with 3895 questions. The scenes cover diverse environments such as urban streets, deserts, forests, ports, and tropical islands, providing a broad test bed for cross-environment generalization. The evaluation procedure for this downstream task is the same as that in Section~3.

Together, SIS-Motion-54K validates motion-centric supervision, while the ablation and downstream studies isolate the contribution of motion-modeling choices and demonstrate transfer to UAV path planning.

\section{Limitations and Future Directions}

Several directions remain open beyond the current evaluation of UAV embodied cognition.

\subsection{Scope and limitations}

The main contribution of SIS-Bench is the explicit factorization of \emph{space} and \emph{self} across perception, memory, and reasoning, rather than entirely new atomic tasks. By standardizing existing datasets and reorganizing established tasks as QA probes, its scope remains limited to these sources, multiple-choice evaluation, and open-loop UAV video understanding. SIS-Motion is a controlled motion-aware probe rather than a highly optimized architecture: on Qwen2.5-VL-3B, it improves overall accuracy from 66.4\% to 69.1\% ($+2.7$ points), but gains remain modest and inconsistent on reasoning tasks. Additional 7B tests yield $+3.8$ and $+5.2$ points for feature-level and LLM-stage motion injection over visual-only SFT, indicating that connector placement matters; broader architectures and closed-loop control remain open.

\subsection{Native multimodal models}

Native multimodal foundation models, including Gemini, Doubao, Qwen3.5, and Kimi, increasingly unify language, vision, motion, and environment feedback. This is particularly relevant to UAV agents, which must interpret dynamic scenes, understand instructions, track ego-motion, and decide under changing viewpoints. Future work should target unified multimodal embodied cognition rather than isolated text reasoning.

\subsection{Integrated sensing}

CLIP-style visual encoders capture coarse semantics but remain limited for fine-grained spatial structure, subtle motion, and long-horizon continuity in UAV videos, especially with small targets, viewpoint changes, altitude variation, and trajectory-dependent reasoning. Future \emph{integrated sensing} should combine appearance, motion, depth, geometry, and temporal correspondence to support stable scene reconstruction and action understanding.

\subsection{Autonomous flying agents}

SIS-Bench evaluates \emph{spatial cognition} and \emph{self-awareness}, and SIS-Motion improves action perception, but question answering remains an intermediate capability. Future autonomous UAV agents should use self-in-space understanding to support navigation, planning, interaction, and long-horizon decision making in open environments, closing the perception--reasoning--action loop.

\end{document}